\documentclass[runningheads]{llncs}
\usepackage{graphicx}
\usepackage{subcaption}
\usepackage{cite}
\usepackage{multicol}
\usepackage{multirow}
\usepackage{amsmath} 
\usepackage{algorithm}
\usepackage{algpseudocode}
\usepackage[hidelinks]{hyperref}
\hypersetup{
  colorlinks   = true, %Colours links instead of ugly boxes
  urlcolor     = blue, %Colour for external hyperlinks
  linkcolor    = blue, %Colour of internal links
  citecolor   = blue %Colour of citations
}
\usepackage{wrapfig}
\usepackage{orcidlink}
\usepackage{subdef}
\begin{document}
\title{\large{\name{}: Script-agnostic Structure Recognition in Tables}}
\titlerunning{\name{}: Script Agnostic TSR}

\author{Dhruv Kudale\inst{1}\inst{*}\orcidlink{0000-0002-3862-819X}\and
Badri Vishal Kasuba\inst{1}\inst{*}\orcidlink{0000-0003-2636-7639}  \and
Venkatapathy Subramanian\inst{1}\orcidlink{0000-0002-4851-628X} \and
Parag Chaudhuri\inst{1}\orcidlink{0000-0002-1706-5703} \and
Ganesh Ramakrishnan\inst{1}\orcidlink{0000-0003-4533-2490}}
\authorrunning{Kudale et al.}

\institute{
\email{ \{dhruvk, badrivishalk, venkatapathy, paragc, ganesh\}@cse.iitb.ac.in} \\
$^1$Department of Computer Science and Engineering, IIT Bombay, Mumbai, India\\
{$^*$These authors contributed equally to this work.}
}

\maketitle

\begin{abstract}
Table Structure Recognition (TSR) is vital for various downstream tasks like information retrieval, table reconstruction, and document understanding. While most state-of-the-art (SOTA) research predominantly focuses on TSR in English documents, the need for similar capabilities in other languages is evident, considering the global diversity of data. Moreover, creating substantial labeled data in non-English languages and training these SOTA models from scratch is costly and time-consuming. We propose TSR as a language-agnostic cell arrangement prediction and introduce \name{} — Script-agnostic Structure Recognition in Tables. \name{} uses recently introduced Optimized Table Structure Language (OTSL) sequences to predict table structures. We show that when coupled with a pre-trained table grid estimator, \name{} can improve the overall tree edit distance-based similarity structure scores of tables even for non-English documents. We experimentally evaluate our performance across benchmark TSR datasets including PubTabNet, FinTabNet, and PubTables-1M. Our findings reveal that \name{} not only matches SOTA models in performance on standard datasets but also demonstrates lower latency. Additionally, \name{} excels in accurately identifying table structures in non-English documents, surpassing current leading models by showing an absolute average increase of 11.12\%.  We also present an algorithm for converting valid OTSL predictions into a widely used HTML-based table representation. To encourage further research, we release our code and Multilingual Scanned and Scene Table Structure Recognition Dataset, \dataset{} labeled with OTSL sequences for 1428 tables in thirteen languages encompassing several scripts at \url{https://github.com/IITB-LEAP-OCR/SPRINT}

\end{abstract}

\section{Introduction}
\begin{figure}[htbp]
    \centering
    \begin{subfigure}[b]{0.48\textwidth}
        \centering
        \includegraphics[width=\textwidth]{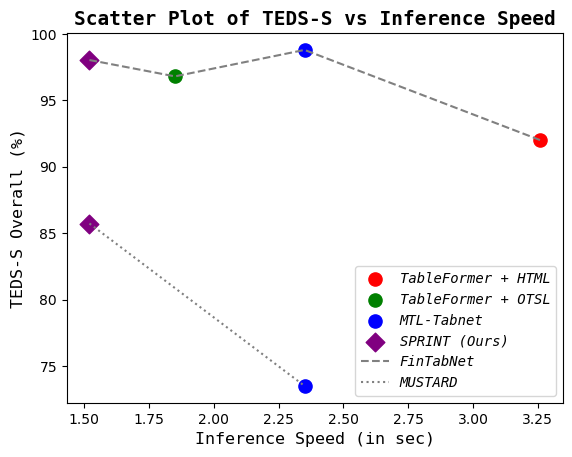}
    \caption{Graph illustrating the comparative TSR performance of \name{} (our approach) along with other approaches \cite{otsl, mtl-tab-net, tableformer} for FinTabNet and \dataset{} datasets.} 
    \label{fig:tsrintro}
    \end{subfigure}%
    \hfill
    \begin{subfigure}[b]{0.48\textwidth}
        \centering
        \begin{subfigure}[b]{\textwidth}
            \centering
            \includegraphics[scale =0.12]{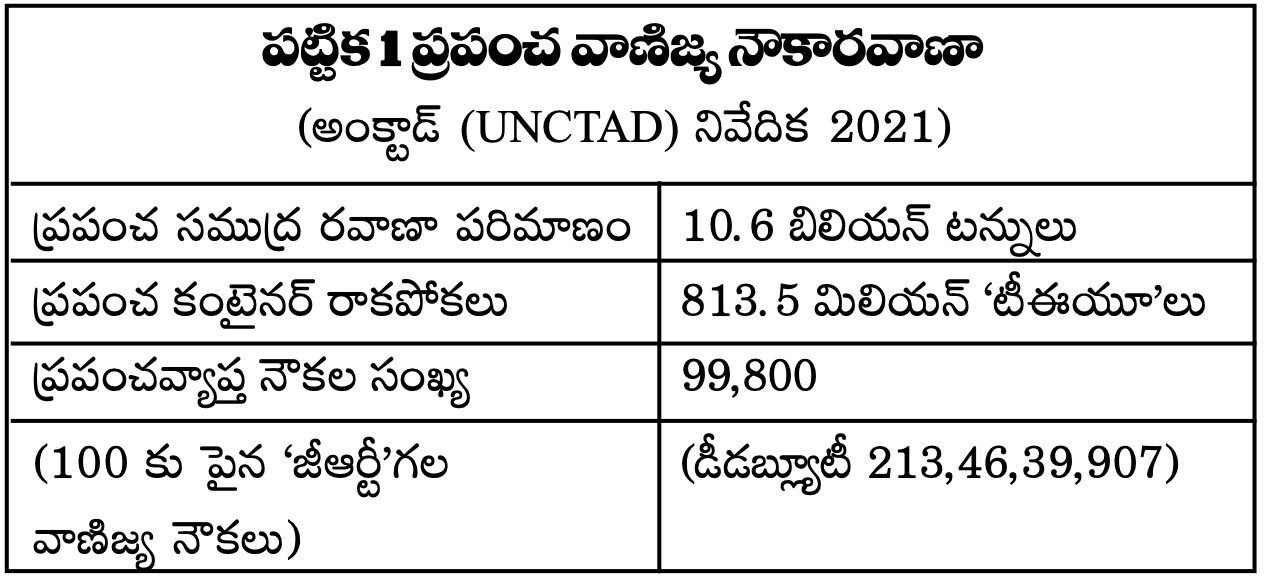}
            \caption{Sample table from \dataset{}}
            \label{fig:tsr-sample}
        \end{subfigure}\\[1ex]
        \begin{subfigure}[b]{0.45\textwidth}
            \centering
            \includegraphics[scale = 0.2]{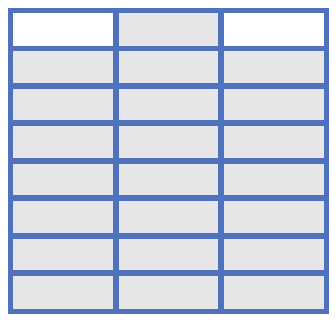}
            \caption{Slow, inaccurate TSR using MTL-TabNet \cite{mtl-tab-net}}
            \label{fig:mtl}
        \end{subfigure}%
        \hfill
        \begin{subfigure}[b]{0.45\textwidth}
            \centering
            \includegraphics[scale=0.2]{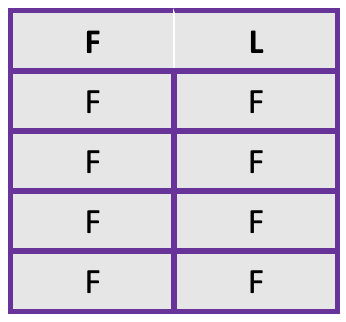}
            \caption{Fast, accurate script agnostic TSR by \name{}}
            \label{fig:sprint}
        \end{subfigure}%
        % \caption{TSR using different approaches}
        \label{fig:right}
    \end{subfigure}
    \caption{Recent performance trends of various Im2Seq models for TSR}
    \label{fig:main-teaser}
\end{figure}

Tables in documents serve as powerful tools for organizing and presenting complex data in a structured and visually comprehensible manner. The structure of tables can vary from simple rows and columns to intricate designs with merged cells and hierarchical arrangements. Additionally, the styling of tables, including multilingual content, font choices, colors, and the presence of borders contribute to the overall diversity of tables. It is necessary to understand the structure of tables to perform OCR, table reconstruction, or any other document analysis tasks. Depending upon the application and nature of representation, table structures are generally divided into two notions namely physical structure and logical structure. It is important to deduce both these structures for various downstream applications. The physical structure of the table is denoted by the actual demarcation of table regions like rows, columns, cells, etc. which are generally represented using bounding boxes. The logical structure of a table represents the underlying topology and gives us more information about the cell adjacency relations. It also conveys more about the cells spanned or merged which benefits table reconstruction in the required format. Logical structures of tables are generally represented using HTML, LATEX, or more recently through OTSL \cite{otsl} sequences. Logical structure prediction is considered to be a sequence generation task. Recently, as seen in Fig \ref{fig:main-teaser}, many Image-to-sequence-based (Im2Seq) models have been employed to predict the logical structures of tables. 

Im2Seq TSR models are transformer-based models that take a table image as an input and produce a sequence denoting the logical structure of the table. A lot of popular Im2Seq models \cite{table-master, mtl-tab-net} have adapted Global Context Attention (GCA) \cite{master} in their encoders to boost the TSR performance on popular benchmark datasets. But even though the usage of such GCA-incorporated Im2Seq models like MTL-TabNet \cite{mtl-tab-net} have improved the TSR performance by proposing new architectures, they suffer from two major drawbacks: First of all, they are unable to generalize well on tables having content in different scripts. Secondly, they end up using 
 only HTML-based tag sequences for representing table structures. Such drawbacks are highlighted in the graph shown in Fig \ref{fig:tsrintro}. MTL-TabNet \cite{mtl-tab-net} that achieves the best TSR performance on the FinTabNet dataset is not only relatively slower but also is unable to generalize well on \dataset{} achieving an overall tree edit distance-based similarity structure (TEDS-S) score of under 75\%. On the other hand, the OTSL-based table structure representation \cite{otsl} has proven to be beneficial for faster decoding. It is also evident that using the same TableFormer \cite{tableformer} architecture, the performance is better and faster using OTSL representation as compared to the HTML representation. Most of these Im2Seq models rely on large HTML tag-based vocabularies making them relatively slower. To design script-agnostic TSR, the model must generalize well on diverse data. \dataset{} comprises of diverse tables from different sources. As shown in Fig \ref{fig:mtl}, MTL-TabNet incorrectly estimates the structure of \dataset{} table shown in Fig \ref{fig:tsr-sample} by interpreting three columns with empty cells on either extreme side of the first row. This is mainly because SOTA models like MTL-TabNet are trained on upsampled table images that mostly have English content and they implicitly capture some language-specific or font-specific features from the image. Since \dataset{} consists of document tables as well as scene tables in several scripts, sizes, and styles, it becomes challenging to perform script-specific or modality-specific TSR. Moreover, all the above-mentioned conventional approaches require training of heavy DL models that require lots of labeled data. Labeled data for tables in non-English languages is limited. To the best of our knowledge, no labeled datasets are available for tables with content in various scripts with their logical structures annotated. 

Thus, we propose \name{}, for a fast and script-agnostic structure recognition in tables that addresses the above highlighted gaps. Since \name{} views table structures as a script-agnostic arrangement of cells, it is beneficial to blur the language-specific or font-specific peculiarities in the table images before they are sent to Im2Seq models. We model the global context of such blurred or downsampled images, using GCA-based encoders as used in SOTA. This helps \name{} to generalize better on tables having content in different scripts. The \name{} decoder also leverages a minimal OTSL vocabulary rather than a larger error-prone HTML vocabulary \cite{otsl}. As seen in Fig \ref{fig:sprint}, the table structure of the sample image (Fig \ref{fig:tsr-sample}) is accurately predicted with two columns and five rows represented using the OTSL sequence. The 'F' in OTSL representation stands for a filled cell and 'L' stands for a left-ward-looking cell indicating a column span of 2 in the first row. Employing OTSL vocabulary helps in faster decoding of structure sequences. Thus, downsampling table images, usage of a GCA-based encoder, and adapting the OTSL-based representation help in making \name{} both fast and script agnostic. Through this work, we make the following contributions 

\begin{enumerate}
    \item We present the architecture and design of \name{}, an Im2Seq model that consists of a GCA-based encoder and a transformer-based decoder that uses the OTSL-based representation for fast and accurate predictions.
    \item We propose the usage of \name{} along with a loosely coupled Table Grid Estimator \cite{tatr-pub-1m} for deducing the logical structure of tables having content in various scripts.
    \item We present an algorithm to convert a well-defined OTSL prediction into an HTML tag-based sequence for accessibility and evaluation purposes. We subsequently highlight the TEDS-S scores achieved by our method for popular TSR datasets. 
    \item Finally we release our code and \dataset{}, a dataset labeled with OTSL-based sequences to represent the logical structures of $1428$ tables having content in several scripts and modalities.
\end{enumerate}

The outline of the rest of the paper is as follows. Section \ref{sec:rel} showcases related work being done in the field of TSR. We describe our methodology in Section \ref{sec:methodology}. Section \ref{sec:expt} gives a detailed overview of our experimentation. We finally present our results in Section \ref{sec:results} and conclude in Section \ref{sec:concl} respectively.

\section{Related Literature}

TSR is a widely studied problem \cite{tsr-survey} encompassing various innovative approaches. TSR methods that determine the complete physical and logical structure of tables can be broadly divided into object detection-based methods and Im2Seq-based methods. 

\textbf{Object detection-based methods} are predominantly used for deducing the physical structure of tables. They detect and demarcate regions in the input table image. These include popular object detection networks like Faster R-CNN \cite{faster-rcnn}, Mask R-CNN \cite{mask-rcnn}, Cascade R-CNN \cite{cascade-rcnn, cascadetabnet}, and YOLO \cite{yolo} to detect cells. Recently transformer-based architectures like DETR \cite{detr} have also been employed for object detection-like tasks that can be used to determine the physical structure of tables. The 'cell bbox decoder' of TableFormer \cite{tableformer}, and TATR \cite{tatr-pub-1m} are some transformer-based approaches that use DETR for determining the cells (as bounding boxes) from detected tables in document images. Various methods like TSRFormer \cite{tsrformer3}, TSR using enhanced DETR \cite{robust-tsr4} and split and merge-based methods \cite{phy1-tsr, phy2-tsr} also follow similar approaches to determine rows and columns of detected tables. Most of the object detection methods predict the physical structure of tables and eventually rely on simple bounding boxes-based post-processing algorithms to associate the detected cells with the logical structure of tables. Hence, these methods can be used in a standalone manner. However, one major drawback of object detection-based methods is their emphasis on enhancing detection performance, which may not necessarily result in a more effective deduction of the logical structure of tables during post-processing. Further, if the object detection stage performs poorly, the deduced structure is also incorrectly estimated. 

\textbf{Im2Seq-based methods} are TSR methods that perform sequence generation from input table images. Transformers have proven to be beneficial for such tasks. Most of these approaches use encoder-decoder-like architectures equipped with attention like the structure decoder of TableFormer \cite{tableformer}, EDD \cite{edd}, etc. TableFormer is an amalgamation of two different models that employ the usage of encoder-decoder with attention \cite{attention} for logical structure and DETR \cite{detr} for physical structure. The output sequence for such Im2Seq models corresponds to a representation of the logical structure of the table. These sequences can be represented in many forms including HTML, Latex, Markdown, etc. Most of these Im2Seq approaches use the HTML representation for decoding the table structure. There are a lot of drawbacks to using the HTML-based vocabulary \cite{otsl} as it can be error-prone and unreasonably large. Recently the Optimized Table Structure Language (OTSL) \cite{otsl} has been introduced to represent tables' logical structure. OTSL is an efficient token-based representation with minimized vocabulary and well-defined rules. Apart from that, the strict syntax and rules to interpret the OTSL structure can also be used to convert it into an HTML sequence. It reduces the number of tokens to be decoded and also helps in faster inference. We exploit this usage of OTSL for predicting the sequence corresponding to the logical structure. Various Im2Seq models like TableMaster \cite{table-master} end up using pre-existing transformer-based models like MASTER \cite{master}. Im2Seq models used for logical structure prediction are seldom standalone. They end up relying on physical structure predictors for tables in documents to assist in the entire reconstruction of tables. 

\textbf{Hybrid, combined, and miscellaneous} works include various new methods like graph-based networks that perform TSR. They usually represent tables as graphs with nodes as table cells, and their edges represent cell relationships by posing this as a graph reconstruction problem. Examples of such approaches include Global Table Extractor (GTE) \cite{gte}, and Table Graph Reconstruction Network (TGRNet) \cite{tgrnet}. Even though graph-based TSR approaches can generalize better, they rely on external graph-based models and require constrained optimizations \cite{graph-based-tsr}. Multi-Type-TD-TSR \cite{tdtsr} proposes a multi-stage pipeline to perform table detection followed by TSR but processes bordered and borderless tables differently. Some other hybrid approaches of TSR include end-to-end attention-based models \cite{tsr-local-attention, mtl-tab-net} that try to predict physical structure, logical structure, and cell content using three or more decoders together. Few approaches try to obtain more aligned bounding boxes (physical structure) by utilizing visual information \cite{lgpma, tabstruct} or reformulating losses (VAST) \cite{vast} to deduce the table structure. Such hybrid works eventually end up working with distinct physical and logical TSR models in synchronization, eliminating the need to explicitly map or align the independently predicted logical and physical structures during the reconstruction stage. 

\textbf{TSR Datasets} include a lot of popular huge TSR datasets like PubTabNet \cite{edd}, FinTabNet \cite{fintabnet}, SynthTabNet \cite{tableformer}, and PubTables-1M \cite{tatr-pub-1m} that have tables from mostly English documents. TabLeX \cite{TabLeX}, SciTSR \cite{scitsr} and TableBank \cite{tablebank} are a few datasets of moderate sizes that focus on document tables. Few older and relatively smaller datasets include ICDAR 2013 \cite{icdar13}, ICDAR 2019 \cite{icdar19}, UNLV \cite{unlv}, etc. All of the datasets mentioned above have data annotated with the logical structure of tables in either HTML or LATEX formats. Recently, a collection of OTSL-based canonical TSR datasets \cite{otsl} has also been released for a subset of tables in the PubTabNet, FinTabNet, and PubTables-1M datasets which we leverage for training \name. TabRecSet \cite{tabrecset} is a recently released bilingual end-to-end table detection, TSR, and table content recognition dataset with tables annotated from English and Chinese documents. However, to the best of our knowledge, there are no multilingual TSR datasets. Taking inspiration from multilingual datasets proposed for text detection \cite{textron} and text recognition \cite{iiit-indic}, we propose \dataset{} having structures annotated for various multilingual tables. Apart from that, a lot of metrics have been used to evaluate TSR models including Mean Average Precision (MAP) for physical structure, TEDS-S score for logical structure, and TEDS score \cite{edd}, GriTS \cite{grits} score for both logical as well as physical structure.
\label{sec:rel}

\section{Our Methodology}
\label{sec:methodology}
\begin{figure}[htbp]
    \centering
    \includegraphics[scale=0.23]{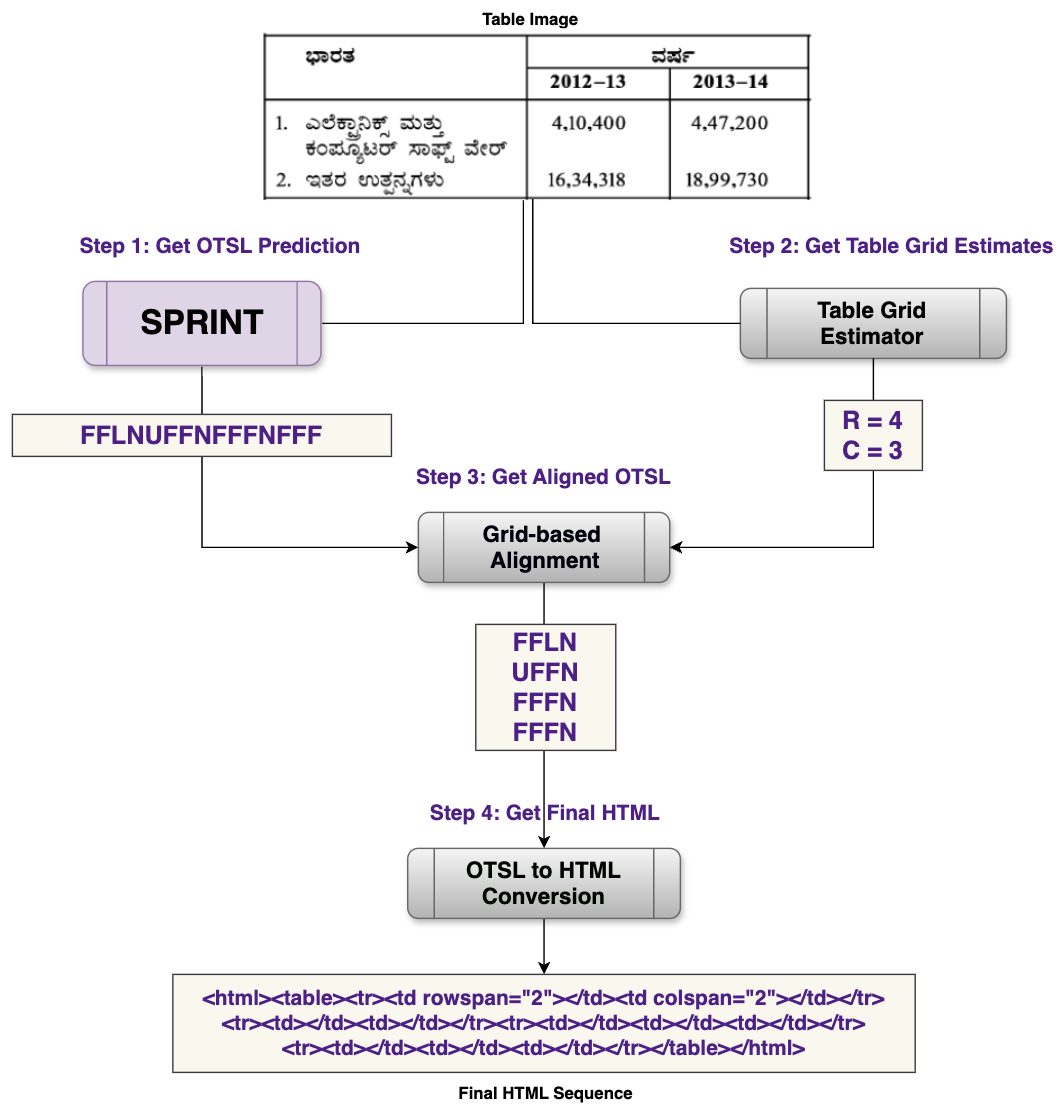}
    \caption{\small{Our methodology for TSR using \name{}}} 
    \label{fig:tsr}
\end{figure}

Our objective is to determine the structure of the input table image. Fig \ref{fig:tsr} highlights the design of our proposed methodology. 
% The input image is provided to \name{} which is our table structure decoder that predicts the sequence of OTSL characters. The same image of the cropped table is provided to a table grid estimator (for physical structure) to get the number of rows and columns. Using the outputs of both \name{} and the table grid estimator we perform grid-based alignment to get a valid OTSL matrix. Finally, this validated OTSL matrix is converted into an HTML tag sequence.  

The first step involves interpreting the logical structure of the input table. This makes use of \name{}, the working and training details of which we explain in Section \ref{expt:crate}. The next step involves interpreting the physical structure of the table by demarcating the rows and columns of the input table image. The goal is to identify the number of objects for two classes namely 'table-row' and 'table-column'. We estimate the number of rows (\row) and number of columns (\col) in this step. The technical details of the table grid estimator are mentioned in Section \ref{expt:grid}. Since OTSL has a well-defined syntax associated with the table structure, it is necessary to convert the predicted string into a valid OTSL matrix. For a table with \row{} rows and \col{} columns, a valid OTSL matrix has \row{} * (\col{} + 1) entries. The estimated number of rows and the number of columns obtained in Step 2 are used to verify the length of the predicted string. The string is made to have a length of \row{} * (\col{} + 1) by appropriate padding or trimming techniques to reshape it into a proper OTSL matrix as shown in the output of Step 3 in Fig \ref{fig:tsr}. The periodicity of character 'N' is also verified by making sure that every (\col{} + 1)$^{th}$ character is 'N'. Similarly, incorrectly placed 'L', 'U', 'X', and 'N' characters are replaced with 'F' denoting a fundamental table cell. This grid-based alignment step is beneficial for two reasons. First of all, it produces a syntactically valid OTSL matrix that can be converted to other popular formats like HTML. Secondly, it eliminates the need to explicitly map the cells of the table with a logical structure through extensive post-processing. As the aligned OTSL sequence and the estimated grid indicate the same number of rows, columns, and cells, every cell is implicitly mapped with a certain row and column by default. Finally, in the last step, the aligned and validated OTSL matrix is converted into an HTML sequence using the procedure mentioned in Algorithm \ref{alg:generate_html_table}. The intuition for the intermediate method for finding the cell spans of a single entry in the OTSL matrix is shown in Algorithm \ref{alg:get_cell_spans}. Eventually, we determine the tree-like representation of the table's structure which can be used for further evaluation and reconstruction. 

\begin{algorithm}
\caption{Generate HTML tag sequence from a valid OTSL matrix}
\label{alg:generate_html_table}
\begin{algorithmic}[1]

\Require $R$ (Number of rows), $C$ (Number of columns), \texttt{otsl\_matrix}: Valid OTSL sequence converted into a $R \times (C+1)$ matrix
\Ensure \texttt{html}: HTML string representing a table

\State \texttt{html} $\gets$ "\textless html\textgreater\textless table\textgreater\textless tbody\textgreater"

\For{$i = 0$ to $R - 1$}
    \State \texttt{html} $\gets$ \texttt{html} + "\textless tr\textgreater"
    \For{$j = 0$ to $C$}
        \State \texttt{entry} $\gets$ \texttt{otsl\_matrix[$i$][$j$]}
        \If{\texttt{entry} is 'E' or 'F'}
            \State ($rs$, $cs$) $\gets$ \texttt{get\_cell\_spans(otsl\_matrix, $i$, $j$)}
            \If{$rs \neq 0$ and $cs \neq 0$}
                \State \texttt{html} $\gets$ \texttt{html} + "\textless td rowspan=\{cs + 1\} colspan=\{rs + 1\}\textgreater\textless/td\textgreater"
            \ElsIf{$rs \neq 0$ and $cs = 0$}
                \State \texttt{html} $\gets$ \texttt{html} + "\textless td colspan=\{rs + 1\}\textgreater\textless/td\textgreater"
            \ElsIf{$rs = 0$ and $cs \neq 0$}
                \State \texttt{html} $\gets$ \texttt{html} + "\textless td rowspan=\{cs + 1\}\textgreater\textless/td\textgreater"
            \Else
                \State \texttt{html} $\gets$ \texttt{html} + "\textless td\textgreater\textless/td\textgreater"
            \EndIf
        \ElsIf{\texttt{entry} is 'N'}
            \State \texttt{html} $\gets$ \texttt{html} + "\textless/tr\textgreater"
        \EndIf
    \EndFor
\EndFor

\State \texttt{html} $\gets$ \texttt{html} + "\textless/tbody\textgreater\textless/table\textgreater\textless/html\textgreater"
\State \Return \texttt{html}

\end{algorithmic}
\end{algorithm}
\begin{algorithm}
\caption{Get cell spans for a corresponding entry in OTSL matrix}
\label{alg:get_cell_spans}
\begin{algorithmic}[1]

\Require \texttt{otsl\_matrix}: Matrix of OTSL sequence, $i$: Row index, $j$: Column index
\Ensure $rs$: Row span, $cs$: Column span

\State \texttt{entry} $\gets$ \texttt{otsl\_matrix[$i$][$j$]}

\If{\texttt{entry} is not 'E' and not 'F'}
    \State \Return $(0, 0)$
\Else
    \State \texttt{row} $\gets$ Character sequence in \texttt{otsl\_matrix[$i$]} from index $j+1$
    \State \texttt{col} $\gets$ Character sequence in column $j$ of \texttt{otsl\_matrix} from index $i+1$
    
    \State $rs$ $\gets$ Count of contiguous occurrences of 'L' in \texttt{row} after \texttt{entry}
    \State $cs$ $\gets$ Count of contiguous occurrences of 'U' in \texttt{col} below \texttt{entry}
    
    \State \Return $(rs, cs)$
\EndIf

\end{algorithmic}
\end{algorithm}

\section{Experiments}
\label{sec:expt}
We begin by mentioning the OTSL-based datasets and then elaborating upon the architecture of \name{} and usage of TATR \cite{tatr-pub-1m} as our table grid estimator that is used to deduce the logical structure of the input table image.

\subsection{Datasets}
\begin{figure}[h]
    \centering
    \makebox[0pt]{\includegraphics[width=\linewidth]{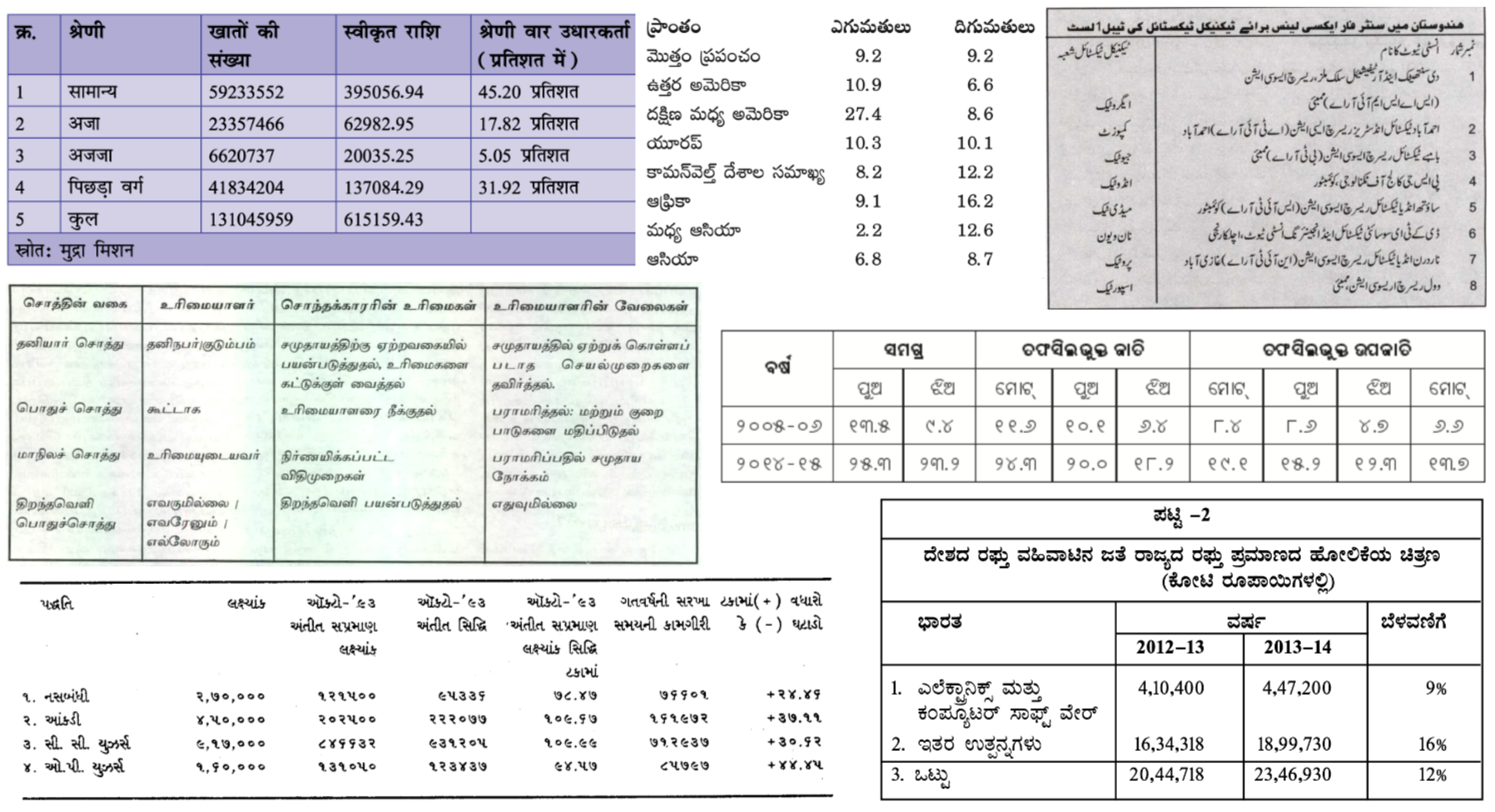}}
    \caption{Sample images from our internal dataset, \dataset{}} 
    \label{fig:indictabnet}
\end{figure}

We have used popular benchmark TSR datasets of PubTabNet \cite{edd}, FinTabNet \cite{fintabnet} and PubTables-1M \cite{tatr-pub-1m} which contain tables predominantly from English documents. The canonical subsets of table images from these datasets have their corresponding OTSL sequences released \cite{otsl} which we use for training and validation purposes. We internally split the train set of PubTabNet for training and validation and further have used the non-overlapping table images in the PubTabNet validation set to report our results to compare our performance with other approaches. Table \ref{tab:datasets} gives a brief idea of the datasets used and the number of images in every split for our experimentation. More details and OTSL-character-specific statistics are mentioned in the supplementary material. To ensure consistency, we use the test sets of canonical datasets to compare with OTSL baselines and that of original datasets to compare with HTML baselines. The further bifurcation of the test sets in terms of the types of tables (simple or complex) is also enlisted. Simple tables do not have any spanned or merged cells, whereas complex tables have at least one merged or spanned cell in them. Besides, we also report results on our internal dataset, \dataset{}. \dataset{} consists of $1428$ tables in thirteen languages that are cropped and annotated from multiple sources \cite{yojana, tabrecset, icdar2019} that originally contain page-level images. \dataset{} encompasses $1214$ cropped document table images (scanned or printed) in twelve languages, including eleven Indian languages \cite{yojana}, each with approximately $100$ tables, namely Assamese, Bengali, Gujarati, Hindi, Kannada, Malayalam, Oriya, Punjabi, Tamil, Telugu, Urdu, and an additional $102$ Chinese cropped tables sourced from CTDAR documents \cite{icdar19, icdar2019}. \dataset{} also includes $214$ English and Chinese scene tables that are cropped and annotated from a subset of images present in the TabRecSet \cite{tabrecset} dataset. A few of the table images from \dataset{} are highlighted in Fig \ref{fig:indictabnet}.

\renewcommand{\arraystretch}{1.1}
\begin{table}[h]
\centering
\resizebox{\textwidth}{!}{
\begin{tabular}{|cc|ccc|cc|}
\hline
\multicolumn{2}{|c|}{\multirow{2}{*}{\textbf{Dataset Name}}} & \multicolumn{3}{c|}{\textbf{Number of Images}}                                                               & \multicolumn{2}{c|}{\textbf{Test Set Details}}     \\ \cline{3-7} 
\multicolumn{2}{|c|}{}                                       & \multicolumn{1}{c|}{\textbf{Training}}           & \multicolumn{1}{c|}{\textbf{Validation}}            & \textbf{Testing} & \multicolumn{1}{c|}{\textbf{Simple}} & \textbf{Complex} \\ \hline
\multicolumn{1}{|c|}{\multirow{2}{*}{PubTabNet \cite{edd}}} & Original  & \multicolumn{1}{c|}{\multirow{2}{*}{320000}} & \multicolumn{1}{c|}{\multirow{2}{*}{68002}} & *9115          & \multicolumn{1}{c|}{4653}            & 4462             \\ \cline{2-2} \cline{5-7} 
\multicolumn{1}{|c|}{}                           & Canonical \cite{otsl} & \multicolumn{1}{c|}{}                         & \multicolumn{1}{c|}{}                        & *6942          & \multicolumn{1}{c|}{4636}            & 2306             \\ \hline
\multicolumn{1}{|c|}{\multirow{2}{*}{FinTabNet \cite{fintabnet}}} & Original  & \multicolumn{1}{c|}{\multirow{2}{*}{88441}}  & \multicolumn{1}{c|}{\multirow{2}{*}{10505}} & 10635         & \multicolumn{1}{c|}{5126}            & 5509             \\ \cline{2-2} \cline{5-7} 
\multicolumn{1}{|c|}{}                           & Canonical \cite{otsl} & \multicolumn{1}{c|}{}                         & \multicolumn{1}{c|}{}                        & 10397         & \multicolumn{1}{c|}{5126}            & 5271             \\ \hline
\multicolumn{2}{|c|}{PubTables-1M Canonical \cite{otsl}}                 & \multicolumn{1}{c|}{522874}                   & \multicolumn{1}{c|}{93989}                   & 92841         & \multicolumn{1}{c|}{44377}           & 48464            \\ \hline
\multicolumn{2}{|c|}{MUSTARD}                                & \multicolumn{1}{c|}{-}                        & \multicolumn{1}{c|}{-}                       & 1428          & \multicolumn{1}{c|}{662}             & 766              \\ \hline
\end{tabular}}
\vspace{3pt}
\caption{Overview of the TSR Datasets used in our experimentation. * indicates evaluating the non-overlapping images in the PubTabNet validation set}
\label{tab:datasets}
\end{table}

\subsection{Image Preprocessing}
\begin{figure}[h]
    \centering
    \begin{subfigure}{0.24\textwidth}
        \includegraphics[width=\linewidth]{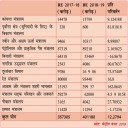}
        \caption{Hindi}
        \label{fig:sub1}
    \end{subfigure}
    \hfill
    \begin{subfigure}{0.24\textwidth}
        \includegraphics[width=\linewidth]{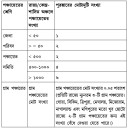}
        \caption{Bengali}
        \label{fig:sub2}
    \end{subfigure}
    \begin{subfigure}{0.24\textwidth}
        \includegraphics[width=\linewidth]{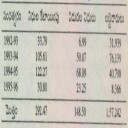}
        \caption{Telugu \& English}
        \label{fig:sub3}
    \end{subfigure}
    \hfill
    \begin{subfigure}{0.24\textwidth}
        \includegraphics[width=\linewidth]{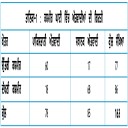}
        \caption{Punjabi}
        \label{fig:sub4}
    \end{subfigure}
    \caption{Resized table images in different languages after preprocessing stage}
    \label{fig:prep}
\end{figure}

We resize all the table images in our datasets to a standard size. Unlike the popular approaches \cite{vast, mtl-tab-net, tableformer}, which upscale the image or pass it to the model as it is, we choose to downsample the images into $128 * 128$ pixels. This preprocessing is done for two main reasons. First of all, it introduces uniformity for the model and generates features of equal dimensions for all images. Secondly, resizing helps to convert the content of table cells in the images into blobs of pixels. These blobs are sufficient enough to convey the presence of some data in the cell and create distortions to ensure that the script-based peculiarities are blurred. This helps to achieve a table image with blobs of pixels representing a script-agnostic arrangement of cells. As seen in Fig \ref{fig:prep}, we see the resized table images originally having content in different languages reduced to a size of $128 * 128$ pixels having blobs. By standardizing these images, \name{} becomes more robust and adaptable to comprehend different tabular structures.

\subsection{Training \name}
\begin{figure}[h]
    \centering
    \makebox[0pt]{\includegraphics[width=\textwidth]{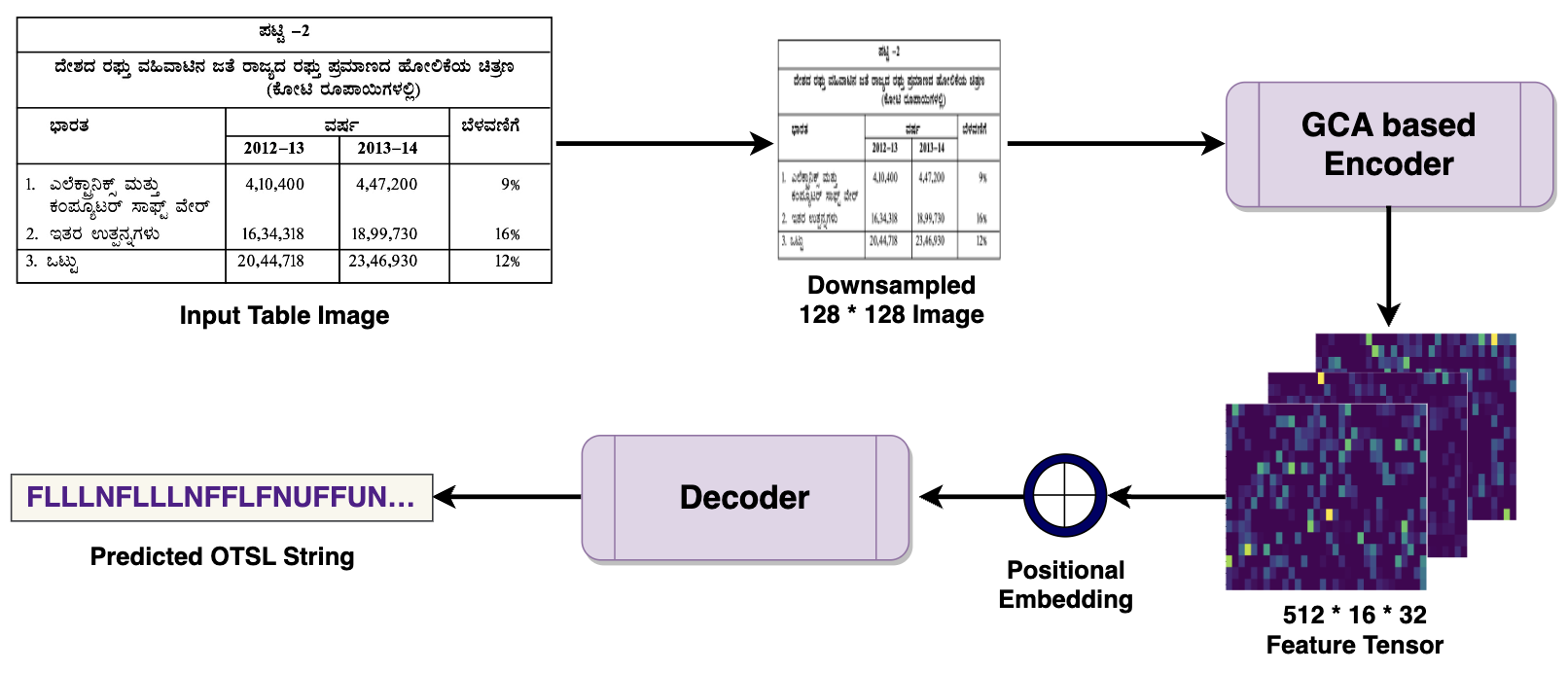}}
    \caption{Architecture diagram explaining the structure and working of \name{}} 
    \label{fig:arch}
\end{figure}

We employ resized images of $128 * 128$ pixels for all subsequent experiments. Our vocabulary comprises six OTSL characters, namely 'F', 'E', 'L', 'X', 'N', and 'U', and \textit{start} and \textit{stop} tokens added manually to the sequences before training. More details about the syntax of representing table structure using OTSL \cite{otsl} sequences and how we have adapted it in \name{} are given in the supplementary material. For every sample, the table image serves as input, with the OTSL sequence as the target (ground truth). As shown in Fig \ref{fig:arch}, \name{} comprises a GCA-based encoder and a transformer-based decoder. We use Multi Aspect Global Attention \cite{master} fused between RESNET31 \cite{resnet} layers in the encoder. The encoder generates a feature tensor that denotes 512 channels of $16 * 32$ feature maps. This feature tensor undergoes positional embedding to generate encoded feature vectors each of size $512$ which are passed on to the decoder. The decoder comprises six layers and a feed-forward neural network with $2048$ nodes in intermediate layers. As most OTSL sequences in our datasets are less than $224$ characters in length, we set the maximum permissible length of predictions as $224$ for the decoder. We use the categorical cross-entropy loss between the ground truths and predicted OTSL sequences to train \name{} using the above-mentioned configuration. One is trained solely on the FinTabNet dataset which we refer to as $\name_{FTN}$. The other model, $\name_{ALL}$, is trained on a merged dataset comprising FinTabNet, PubTabNet, and PubTables-1M. Both models undergo training for over $80$ epochs with a learning rate of $0.0001$. We have performed all our training and relevant experiments on an NVIDIA RTX A6000 single GPU.

% , with the $\name_{FTN}$ requiring approximately 20 minutes per epoch and the $\name_{ALL}$ taking around 135 minutes per epoch. 

\label{expt:crate}

\subsection{Table Grid Estimator}
We have used the TATR \cite{aligning} V1.1 model pretrained on FinTabNet, PubTabNet, and PubTables-1M datasets for determining the physical structure of the input table image. The underlying architecture for TATR is DETR \cite{detr} which is trained for six object classes. For grid alignment, we make use of two classes of 'table-row' and 'table-column' as they are sufficient to accurately determine the number of rows (\row) and columns (\col). We use a detection threshold of $0.25$ during the inference of the TATR V1.1 model. We further carry out NMS for table row class with an IOU threshold of $0.25$ to minimize overlapping predictions and improve the row match. Once \row{} and \col{} are determined, these values align with the OTSL sequence predicted by \name. The supplementary material shows the qualitative results of the TATR model for detecting table rows and table columns respectively. Eventually, it is the number of detections that are used to determine the number of rows (\row) and columns (\col) to align the OTSL sequence. The actual bounding boxes are useful only for further reconstruction of tables. However, only \row{} and \col{} values are sufficient to align the OTSL sequence and deduce the logical structure. To show that, pretrained TATR is a promising grid estimator to determine the number of rows and columns, we present the exact match (in \%) and mean L1 error in estimating the number of rows, number of columns, and both rows and columns together for the different test sets in Table \ref{tab:phy-match}. As shown in the table, for all datasets under consideration, there is a highly accurate match for PubTabNet, FinTabNet, and PubTables-1M datasets. There is a decent match of rows and columns for \dataset{} because TATR is pretrained from abundant tables from only English documents. Even though the match in the number of rows and columns for \dataset{} has scope for improvement, the average L1 error in predicted numbers and actual numbers is less than $0.55$ indicating that the difference in the predicted and actual rows is quite low in number. This low L1 error ensures that the TEDS-S score is not penalized much for the HTML sequence generated which is evident in our results in Section \ref{sec:results}. The better the exact match accuracy, the better would be the aligned OTSL and thus more accurately would be the table structure preserved.
\label{expt:grid}

\renewcommand{\arraystretch}{1.1}
\begin{table}[h]
\centering
\resizebox{0.85\textwidth}{!}{
\begin{tabular}{|c|cc|cc|cc|}
\hline
\multirow{2}{*}{\textbf{Test Dataset}} &
  \multicolumn{2}{c|}{\textbf{Rows Only}} &
  \multicolumn{2}{c|}{\textbf{Columns Only}} &
  \multicolumn{2}{c|}{\textbf{Rows and Cols}} \\ \cline{2-7} 
 &
  \multicolumn{1}{c|}{\textbf{\begin{tabular}[c]{@{}c@{}}Exact \\ Match\end{tabular}}} &
  \textbf{\begin{tabular}[c]{@{}c@{}}Avg L1 \\ Error\end{tabular}} &
  \multicolumn{1}{c|}{\textbf{\begin{tabular}[c]{@{}c@{}}Exact \\ Match\end{tabular}}} &
  \textbf{\begin{tabular}[c]{@{}c@{}}Avg L1 \\ Error\end{tabular}} &
  \multicolumn{1}{c|}{\textbf{\begin{tabular}[c]{@{}c@{}}Exact \\ Match\end{tabular}}} &
  \textbf{\begin{tabular}[c]{@{}c@{}}Avg L1 \\ Error\end{tabular}} \\ \hline
PubTabNet  &
  \multicolumn{1}{c|}{88.82} &
  0.278 &
  \multicolumn{1}{c|}{90.97} &
  0.138 &
  \multicolumn{1}{c|}{81.28} &
  0.208 \\ \hline
FinTabNet  &
  \multicolumn{1}{c|}{88.51} &
  0.212 &
  \multicolumn{1}{c|}{98.42} &
  0.019 &
  \multicolumn{1}{c|}{87.22} &
  0.115 \\ \hline
PubTables-1M  &
  \multicolumn{1}{c|}{94.39} &
  0.123 &
  \multicolumn{1}{c|}{98.19} &
  0.022 &
  \multicolumn{1}{c|}{92.81} &
  0.073 \\ \hline
MUSTARD &
  \multicolumn{1}{c|}{67.67} &
  0.576 &
  \multicolumn{1}{c|}{75.00} &
  0.451 &
  \multicolumn{1}{c|}{54.62} &
  0.513 \\ \hline
\end{tabular}}
\caption{Performance of TATR \cite{tatr-pub-1m} as our table grid estimator for determining the number of rows and columns in the image. We use v1.1-PubTables-1m for reporting results on the PubTables-1M dataset and v1.1-all on the other datasets.}
\label{tab:phy-match}
\end{table}

\section{Results and Discussions}
\label{sec:results}
In this section, we present the TEDS-S scores achieved by \name{} along with the assistance of the table grid estimator. The final output is an HTML tag sequence. PubTabNet, FinTabNet, and PubTables-1M have their ground truths available in HTML format. To evaluate only the structure, we filter the content and only consider the pure HTML tag sequence as the corresponding ground truths. The OTSL-labeled sequences of the \dataset{} are converted to HTML tag sequences using Algorithm \ref{alg:generate_html_table}.

\subsection{Performance on TSR in English Documents}
Table \ref{tab:otsl} highlights our performance against the pre-reported TEDS-S scores of the OTSL baseline \cite{otsl} for all three datasets. We have reported our results on the canonical test sets of FinTabNet, PubTabNet, and PubTables-1M \cite{otsl}. We have used $\name_{FTN}$ to evaluate the FinTabNet canonical test set and $\name_{ALL}$ on the other two test sets respectively. We consistently perform better than the OTSL baseline for all the enlisted datasets. We compare our performance with the pre-reported TEDS-S scores of different approaches that use the HTML-based vocabulary for TSR tasks in Table \ref{tab:html}. We use $\name_{FTN}$ on FinTabNet and $\name_{ALL}$ on the PubTabNet test set for evaluation. We match the performance of the best performing MTL-TabNet \cite{mtl-tab-net} for both datasets. Our approach falls short by an average of 1.5\% because both VAST and MTL-TabNet rely on upscaling images leading to larger feature maps that are leveraged by cascaded decoders trained on the HTML-based vocabulary. Further, MTL-TabNet uses two decoders for decoding the table structure making the model slower than our approach. After testing for 50 iterations on a random subset of 400 FinTabNet images, it is observed that while MTL-TabNet takes an average of \textit{2.35 seconds} per prediction, \name{} can give faster prediction in an average time of \textit{1.52 seconds} per image indicating better latency. We give a detailed overview of inference time estimation in the supplementary material. We are faster not only due to the small-sized feature tensors produced by downsampled images by the GCA-based encoder \cite{master} but also due to the usage of minimized OTSL vocabulary \cite{otsl} for the decoder to produce faster predictions. Besides, our approach also generalizes well for tables having content in other languages (scripts) as described in the upcoming Section \ref{sec:mustard}. 

\renewcommand{\arraystretch}{1.1}
\begin{table}[h]
\centering
\resizebox{\textwidth}{!}{
\begin{tabular}{|c|ccc|ccc|}
\hline
\multirow{2}{*}{\textbf{Dataset}} &
  \multicolumn{3}{c|}{\textbf{TableFormer + OTSL \cite{otsl}}} &
  \multicolumn{3}{c|}{\textbf{Ours}} \\ \cline{2-7} 
 &
  \multicolumn{1}{c|}{\textbf{\begin{tabular}[c]{@{}c@{}}TEDS-S\\ Simple\end{tabular}}} &
  \multicolumn{1}{c|}{\textbf{\begin{tabular}[c]{@{}c@{}}TEDS-S \\ Complex\end{tabular}}} &
  \textbf{\begin{tabular}[c]{@{}c@{}}TEDS-S\\ Overall\end{tabular}} &
  \multicolumn{1}{c|}{\textbf{\begin{tabular}[c]{@{}c@{}}TEDS-S\\ Simple\end{tabular}}} &
  \multicolumn{1}{c|}{\textbf{\begin{tabular}[c]{@{}c@{}}TEDS-S \\ Complex\end{tabular}}} &
  \textbf{\begin{tabular}[c]{@{}c@{}}TEDS-S\\ Overall\end{tabular}} \\ \hline
PubTabNet &
  \multicolumn{1}{c|}{96.50} &
  \multicolumn{1}{c|}{93.40} &
  95.50 &
  \multicolumn{1}{c|}{\textbf{98.20}} &
  \multicolumn{1}{c|}{\textbf{96.24}} &
  \textbf{97.55} \\ \hline
FinTabNet &
  \multicolumn{1}{c|}{95.50} &
  \multicolumn{1}{c|}{96.10} &
  95.90 &
  \multicolumn{1}{c|}{\textbf{98.36}} &
  \multicolumn{1}{c|}{\textbf{97.99}} &
  \textbf{98.17} \\ \hline
PubTables-1M &
  \multicolumn{1}{c|}{98.70} &
  \multicolumn{1}{c|}{96.40} &
  \textbf{*97.70} &
  \multicolumn{1}{c|}{\textbf{98.92}} &
  \multicolumn{1}{c|}{\textbf{96.54}} &
  \textbf{97.68} \\ \hline
\end{tabular}}
\caption{Comparison of our approach against the OTSL baseline on popular TSR datasets for tables in English documents. * indicates that OTSL \cite{otsl} have reported percentage score rounded off to one digit after decimal point}
\label{tab:otsl}
\end{table}
\noindent

\renewcommand{\arraystretch}{1.1}
\begin{table}[]
\centering
\resizebox{\textwidth}{!}{
\begin{tabular}{|c|ccc|ccc|}
\hline
\textbf{Dataset} &
  \multicolumn{3}{c|}{\textbf{PubTabNet \cite{edd}}} &
  \multicolumn{3}{c|}{\textbf{FinTabNet \cite{fintabnet}}} \\ \hline
\textbf{Model} &
  \multicolumn{1}{c|}{\textbf{\begin{tabular}[c]{@{}c@{}}TEDS-S\\ Simple\end{tabular}}} &
  \multicolumn{1}{c|}{\textbf{\begin{tabular}[c]{@{}c@{}}TEDS-S\\ Complex\end{tabular}}} &
  \textbf{\begin{tabular}[c]{@{}c@{}}TEDS-S\\ Overall\end{tabular}} &
  \multicolumn{1}{c|}{\textbf{\begin{tabular}[c]{@{}c@{}}TEDS-S\\ Simple\end{tabular}}} &
  \multicolumn{1}{c|}{\textbf{\begin{tabular}[c]{@{}c@{}}TEDS-S \\ Complex\end{tabular}}} &
  \textbf{\begin{tabular}[c]{@{}c@{}}TEDS-S\\ Overall\end{tabular}} \\ \hline
EDD \cite{edd, rethinking-det} &
  \multicolumn{1}{c|}{91.10} &
  \multicolumn{1}{c|}{88.70} &
  89.90 &
  \multicolumn{1}{c|}{88.40} &
  \multicolumn{1}{c|}{92.08} &
  90.60 \\ \hline
GTE \cite{gte} &
  \multicolumn{1}{c|}{-} &
  \multicolumn{1}{c|}{-} &
  93.01 &
  \multicolumn{1}{c|}{-} &
  \multicolumn{1}{c|}{-} &
  91.02 \\ \hline
TableFormer \cite{tableformer} &
  \multicolumn{1}{c|}{98.50} &
  \multicolumn{1}{c|}{95.00} &
  96.75 &
  \multicolumn{1}{c|}{97.50} &
  \multicolumn{1}{c|}{96.00} &
  96.80 \\ \hline
VAST \cite{vast} &
  \multicolumn{1}{c|}{-} &
  \multicolumn{1}{c|}{-} &
  97.23 &
  \multicolumn{1}{c|}{-} &
  \multicolumn{1}{c|}{-} &
  98.63 \\ \hline
MTL-TabNet \cite{mtl-tab-net} &
  \multicolumn{1}{c|}{\textbf{99.05}} &
  \multicolumn{1}{c|}{\textbf{96.66}} &
  \textbf{97.88} &
  \multicolumn{1}{c|}{\textbf{99.07}} &
  \multicolumn{1}{c|}{\textbf{98.46}} &
  \textbf{98.79} \\ \hline
\textbf{Ours} &
  \multicolumn{1}{c|}{98.00} &
  \multicolumn{1}{c|}{93.32} &
  95.71 &
  \multicolumn{1}{c|}{98.35} &
  \multicolumn{1}{c|}{97.74} &
  98.03 \\ \hline
\end{tabular}}
\caption{Comparison of our approach on PubTabNet and FinTabNet datasets for tables in English documents with other pre-reported TEDS-S scores of HTML-based Im2Seq approaches}
\label{tab:html}
\end{table}

\noindent

\subsection{Performance on \dataset}
\begin{figure}[!h]
    \centering
    \makebox[0pt]{\includegraphics[scale=0.16]{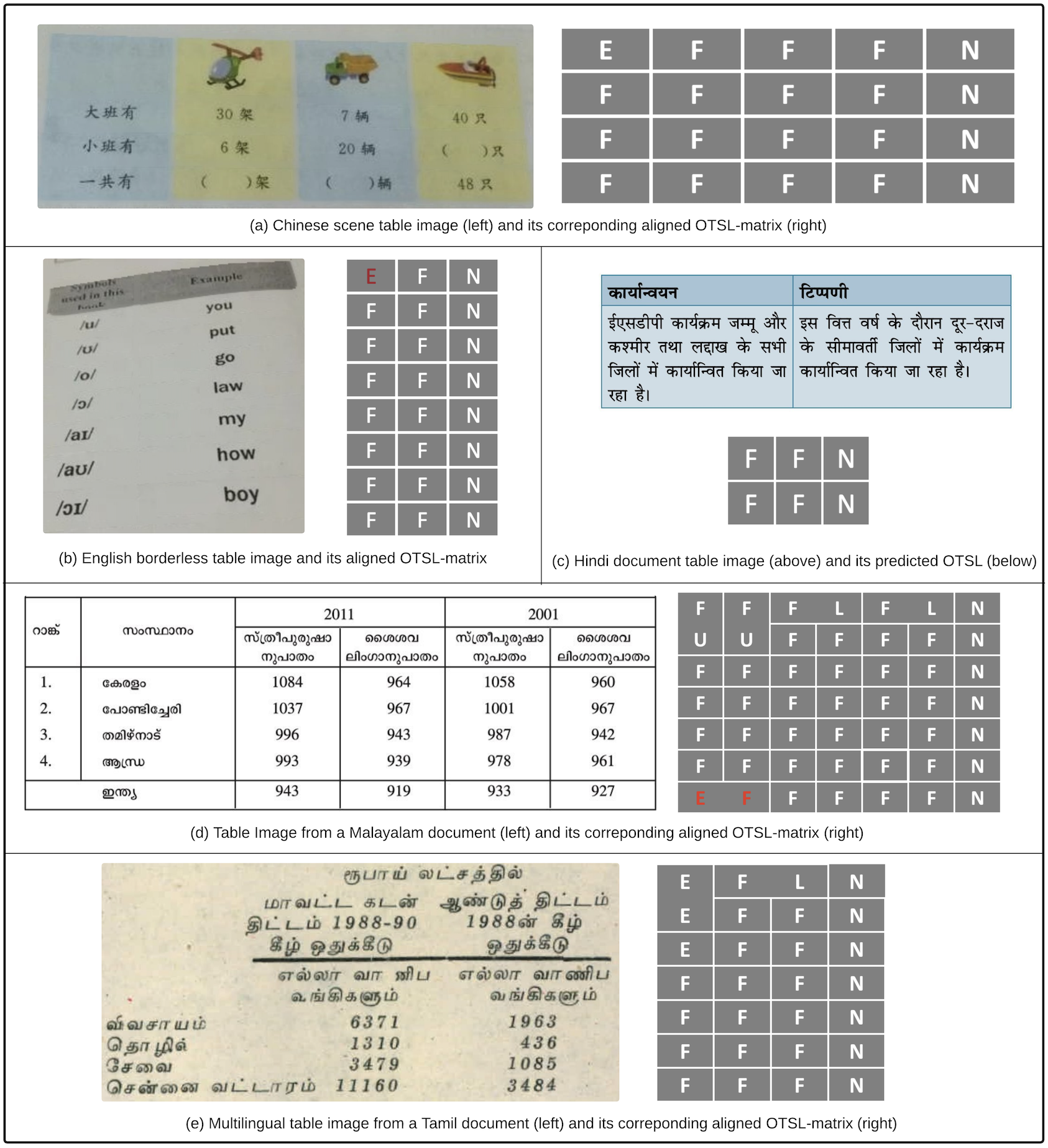}}
    \caption{Each subfigure showcases the final aligned OTSL matrix predicted by $\name_{FTN}$ adjacent to the input table image. The OTSL matrix is projected on an HTML-based table skeleton reflecting the ground truth structure. The red characters denote incorrect OTSL prediction that alters TEDS-S score} 
    \label{fig:qual}
\end{figure}

Table \ref{tab:indic} shows the comparative overview of our performance against MTL-TabNet \cite{mtl-tab-net}. Since source code and checkpoints for OTSL baselines \cite{otsl} and VAST \cite{vast} have not been released, we compare the TEDS-S scores obtained by our approach with the MTL-TabNet (SOTA) scores. For determining TEDS-S on \dataset{}, we use the MTL-TabNet checkpoint trained on the FinTabNet dataset and $\name_{FTN}$ for evaluating our approach. Our approach consistently performs better than MTL-TabNet for all the enlisted languages and shows an average increase of \textit{11.12\%} in the overall TEDS-S score. We believe that MTL-TabNet does not perform as well as our approach since it has been extensively trained on English TSR datasets for upsampled images on HTML vocabulary. As a result, MTL-TabNet is unable to capture the script-agnostic arrangement of cells. Fig \ref{fig:qual} showcases our results on a few images from \dataset{} which show our aligned OTSL predictions projected on the HTML-based table skeleton reflecting the ground truth structures. 
\label{sec:mustard}

\renewcommand{\arraystretch}{1.1}
\begin{table}[h]
\centering
\resizebox{\textwidth}{!}{
\begin{tabular}{|cc|ccc|ccc|}
\hline
\multicolumn{1}{|c|}{\multirow{2}{*}{\textbf{Modality}}} &
  \multirow{2}{*}{\textbf{Language}} &
  \multicolumn{3}{c|}{\textbf{MTL-TabNet}} &
  \multicolumn{3}{c|}{\textbf{Ours}} \\ \cline{3-8} 
\multicolumn{1}{|c|}{} &
   &
  \multicolumn{1}{c|}{\textbf{\begin{tabular}[c]{@{}c@{}}TEDS-S\\ Simple\end{tabular}}} &
  \multicolumn{1}{c|}{\textbf{\begin{tabular}[c]{@{}c@{}}TEDS-S\\ Complex\end{tabular}}} &
  \textbf{\begin{tabular}[c]{@{}c@{}}TEDS-S\\ Overall\end{tabular}} &
  \multicolumn{1}{c|}{\textbf{\begin{tabular}[c]{@{}c@{}}TEDS-S\\ Simple\end{tabular}}} &
  \multicolumn{1}{c|}{\textbf{\begin{tabular}[c]{@{}c@{}}TEDS-S\\ Complex\end{tabular}}} &
  \textbf{\begin{tabular}[c]{@{}c@{}}TEDS-S\\ Overall\end{tabular}} \\ \hline
\multicolumn{1}{|c|}{\multirow{12}{*}{\begin{tabular}[c]{@{}c@{}}Document \\ Tables\\ (Printed and \\ Scanned)\end{tabular}}} &
  Assamese &
  \multicolumn{1}{c|}{79.39} &
  \multicolumn{1}{c|}{73.40} &
  76.54 &
  \multicolumn{1}{c|}{\textbf{88.09}} &
  \multicolumn{1}{c|}{\textbf{88.74}} &
  \textbf{88.40} \\ \cline{2-8} 
\multicolumn{1}{|c|}{} &
  Bengali &
  \multicolumn{1}{c|}{71.68} &
  \multicolumn{1}{c|}{60.02} &
  61.42 &
  \multicolumn{1}{c|}{\textbf{77.24}} &
  \multicolumn{1}{c|}{\textbf{78.52}} &
  \textbf{78.36} \\ \cline{2-8} 
\multicolumn{1}{|c|}{} &
  Gujarati &
  \multicolumn{1}{c|}{85.12} &
  \multicolumn{1}{c|}{76.72} &
  79.63 &
  \multicolumn{1}{c|}{\textbf{87.79}} &
  \multicolumn{1}{c|}{\textbf{81.34}} &
  \textbf{83.58} \\ \cline{2-8} 
\multicolumn{1}{|c|}{} &
  Hindi &
  \multicolumn{1}{c|}{73.80} &
  \multicolumn{1}{c|}{76.60} &
  75.04 &
  \multicolumn{1}{c|}{\textbf{85.68}} &
  \multicolumn{1}{c|}{\textbf{88.22}} &
  \textbf{86.81} \\ \cline{2-8} 
\multicolumn{1}{|c|}{} &
  Kannada &
  \multicolumn{1}{c|}{68.82} &
  \multicolumn{1}{c|}{66.73} &
  67.20 &
  \multicolumn{1}{c|}{\textbf{71.84}} &
  \multicolumn{1}{c|}{\textbf{79.02}} &
  \textbf{77.34} \\ \cline{2-8} 
\multicolumn{1}{|c|}{} &
  Malayalam &
  \multicolumn{1}{c|}{82.57} &
  \multicolumn{1}{c|}{79.34} &
  81.07 &
  \multicolumn{1}{c|}{\textbf{86.41}} &
  \multicolumn{1}{c|}{\textbf{85.13}} &
  \textbf{85.81} \\ \cline{2-8} 
\multicolumn{1}{|c|}{} &
  Oriya &
  \multicolumn{1}{c|}{85.28} &
  \multicolumn{1}{c|}{78.03} &
  82.84 &
  \multicolumn{1}{c|}{\textbf{91.55}} &
  \multicolumn{1}{c|}{\textbf{85.20}} &
  \textbf{89.41} \\ \cline{2-8} 
\multicolumn{1}{|c|}{} &
  Punjabi &
  \multicolumn{1}{c|}{65.08} &
  \multicolumn{1}{c|}{48.63} &
  51.54 &
  \multicolumn{1}{c|}{\textbf{86.91}} &
  \multicolumn{1}{c|}{\textbf{79.65}} &
  \textbf{80.93} \\ \cline{2-8} 
\multicolumn{1}{|c|}{} &
  Tamil &
  \multicolumn{1}{c|}{81.96} &
  \multicolumn{1}{c|}{71.88} &
  77.83 &
  \multicolumn{1}{c|}{\textbf{94.91}} &
  \multicolumn{1}{c|}{\textbf{85.87}} &
  \textbf{91.21} \\ \cline{2-8} 
\multicolumn{1}{|c|}{} &
  Telugu &
  \multicolumn{1}{c|}{85.07} &
  \multicolumn{1}{c|}{79.28} &
  82.17 &
  \multicolumn{1}{c|}{\textbf{93.70}} &
  \multicolumn{1}{c|}{\textbf{86.00}} &
  \textbf{89.85} \\ \cline{2-8} 
\multicolumn{1}{|c|}{} &
  Urdu &
  \multicolumn{1}{c|}{70.94} &
  \multicolumn{1}{c|}{69.74} &
  70.03 &
  \multicolumn{1}{c|}{\textbf{81.39}} &
  \multicolumn{1}{c|}{\textbf{75.38}} &
  \textbf{76.86} \\ \cline{2-8} 
\multicolumn{1}{|c|}{} &
  Chinese &
  \multicolumn{1}{c|}{92.43} &
  \multicolumn{1}{c|}{81.58} &
  86.15 &
  \multicolumn{1}{c|}{\textbf{98.11}} &
  \multicolumn{1}{c|}{\textbf{86.00}} &
  \textbf{91.10} \\ \hline
\multicolumn{1}{|c|}{\multirow{2}{*}{\begin{tabular}[c]{@{}c@{}}Scene \\ Tables\end{tabular}}} &
  English &
  \multicolumn{1}{c|}{76.19} &
  \multicolumn{1}{c|}{78.01} &
  76.53 &
  \multicolumn{1}{c|}{\textbf{88.98}} &
  \multicolumn{1}{c|}{\textbf{76.14}} &
  \textbf{85.71} \\ \cline{2-8} 
\multicolumn{1}{|c|}{} &
  Chinese &
  \multicolumn{1}{c|}{69.40} &
  \multicolumn{1}{c|}{66.65} &
  68.94 &
  \multicolumn{1}{c|}{\textbf{88.62}} &
  \multicolumn{1}{c|}{\textbf{81.96}} &
  \textbf{87.27} \\ \hline
\multicolumn{2}{|c|}{Overall} &
  \multicolumn{1}{c|}{77.70} &
  \multicolumn{1}{c|}{71.90} &
  74.07 &
  \multicolumn{1}{c|}{\textbf{87.23}} &
  \multicolumn{1}{c|}{\textbf{82.66}} &
  \textbf{85.19} \\ \hline
\end{tabular}}
\caption{Comparing MTL-TabNet \cite{mtl-tab-net} with our approach on tables in \dataset{} for various languages (scripts) and modalities}
\label{tab:indic}
\end{table}
\noindent

\section{Conclusion}
\label{sec:concl}

We present \name{} as a solution for fast, robust and script-agnostic TSR. It achieves this via the downsampling of input images, GCA-based encoder, and OTSL-based table structure representation. The minimized OTSL vocabulary not only helps in faster decoding but also in conversions to more popular formats like HTML making the scope of this task more extensive and versatile. We also present an algorithm to convert the OTSL sequence to an HTML-based representation of the table structure. Finally, we release \dataset{} which has been meticulously annotated using OTSL sequences, opening new doors for further research in script-agnostic TSR. In the future, we want to investigate incorporating cells detected using the table grid estimator with the predicted logical structure, to reconstruct complex tables in documents as accurately as possible. Along with the help of table detection models and powerful OCR-based frameworks, we can design end-to-end pipelines that reconstruct the structure and text inside a wide variety of tables having content in different scripts.

\subsubsection*{Acknowledgement}
We acknowledge the support of a grant from IRCC, IIT Bombay, and
MEITY, Government of India, through the National Language Translation Mission-Bhashini project.

% \bibliographystyle{splncs04}
% \bibliography{main}

\author{}
\title{\large{\name{}: Script-agnostic Structure Recognition in Tables \\ Supplementary Material}}
\institute{}
\titlerunning{\name{}: Script Agnostic TSR}
\authorrunning{Kudale et al.}
\maketitle

We highlight some background knowledge and technical specifications necessary to understand the structure and functioning of \name{}. The following sections give a detailed overview of the notions of table structures, interpretation of OTSL vocabulary, and statistics of the datasets used in our experimentation. Further, we also highlight an illustrative example and give a brief formulation of our inference time calculation. 

\section{Understanding Table Structures}
Depending upon the application and nature of representation, table structures are generally divided into two notions namely physical structure (for retrieval of content) and logical structure (for reconstruction purposes). Both the logical and physical structures are important to deduce for various downstream applications. 

\subsection{Physical Structure}
The physical structure of the table is denoted by the actual demarcation of table regions like rows, columns, cells, etc. in the table image. These regions are generally represented using bounding boxes. These bounding boxes are responsible for retrieving the textual content present in them using suitable OCR engines. As seen in Figure \ref{fig:tsrintro} from TableFormer \cite{tableformer}, the set of red bounding boxes denoting table cells constitute the physical structure of the table image. Physical structure prediction is generally viewed as an object detection task. 

\begin{figure}
    \centering
    \includegraphics[width=\textwidth]{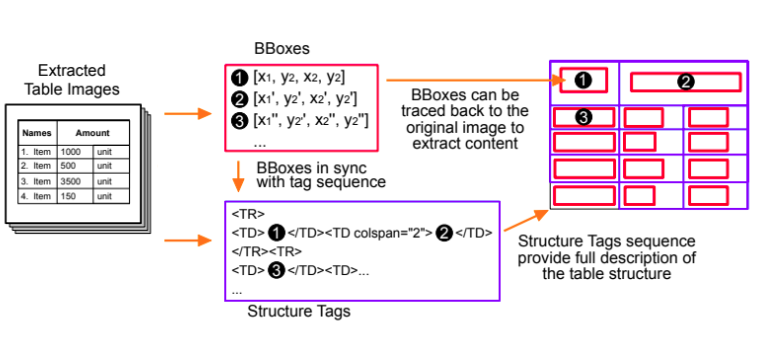}
    \caption{Understanding Table Structure Recognition (TSR) with notions of physical and logical structures as illustrated in TableFormer \cite{tableformer}} 
    \label{fig:tsrintro}
\end{figure}

\subsection{Logical Structure}
The logical structure of a table represents the underlying topology. It gives us more information about the number of rows, columns, and cell adjacency relations. It also conveys more about the cells spanned or merged which in turn proves to be beneficial for reconstructing the table in the required format. As seen in Figure \ref{fig:tsrintro}, the blue skeletal figure on the right represents the logical structure of the table. Logical structures of tables are generally represented using HTML or LATEX sequence of tags. Logical structure prediction of a table is considered to be a sequence generation task.

\section{Understanding OTSL Representation}
Optimized Table Structure Language (OTSL) \cite{otsl} is a recently introduced table structure representation. OTSL is one of the many ways to represent table topologies. Following are a few of its salient features which highlight the benefits of its usage. 

\begin{itemize}
  \item \textbf{Efficient Token-Based Representation:} OTSL utilizes a minimized vocabulary of tokens of five letters (or six if we represent empty cells with another token) to represent the logical structure of tables.
  \item \textbf{Well-Defined Rules:} OTSL has clear rules governing how tokens are structured and combined, ensuring consistency and predictability. This also helps in error detection and resolution of predicted sequences.
  \item \textbf{Conversion to HTML:} An OTSL string can be converted into widely-used HTML sequences preserving all the encoded information. An independent conversion of a valid OTSL string into an HTML tag sequence is always lossless \cite{otsl}. This encourages interoperability and also helps to evaluate tree-edit distance-based similarity metrics.
  \item \textbf{Faster Inference:} Due to a reasonably small vocabulary of six instead of a size of more than 30 in the case of HTML representation, it is beneficial to train an Im2Seq model on the OTSL sequences. This facilitates faster inference by reducing the complexity of the representation of decoder logits and aiding in quicker processing.
\end{itemize}

The OTSL vocabulary \cite{otsl} used in our experimentation consists of six characters namely 'F', 'E', 'L', 'X', 'N', and 'U'. Following is the interpretation of each of the characters:
\begin{itemize}
  \item \textbf{``F'' cell:} indicates a fundamental table cell that has some content.
  \item \textbf{``E'' cell:} indicates a table cell that has no cell content.
  \item \textbf{``L'' cell:} indicates a left-looking cell. This means that it merges with its left neighbor cell and creates a column span.
  \item \textbf{``X'' cell:} indicates a cross cell that merges with both left and upper neighbor cells. This is responsible for adding both row span and column span.
  \item \textbf{``N'':} This token indicates a new row or new line. This means that the parser needs to switch to the next row of the table.
  \item \textbf{``U'' cell:} indicates an upward-looking cell that merges with the upper neighbor cell to create a row span.
\end{itemize}

We train \name{} on these OTSL-based sequences with a vocabulary of above six characters along with special \textit{start} and \textit{stop} characters. We perform validation checks and make appropriate modifications to make the predicted OTSL abide by all syntax rules \cite{otsl} in the grid alignment step of our methodology.

\section{Dataset Details}
The dataset that we release, \dataset{}, comprises tables in multiple languages (scripts) having scanned, printed, or scene-text modalities. The tables have been collected from multiple sources. The scene-text tables of Chinese and English languages have been accumulated from the TabRecSet dataset\cite{tabrecset} covering both bordered and borderless tables. The original TabRecSet is a page-level dataset, so we have cropped and annotated the table-level images and included about 200 images in \dataset. The printed tables of eleven Indian languages have been collected through archives of monthly magazines \cite{yojana} spanning the 1970s to 2020s time period. The tables in the magazine document pages have been cropped and have been annotated by various annotators. For the tables in the printed modality in the Chinese language, we have cropped images from the CTDAR dataset \cite{icdar19} of which we have collected 102 tables. The modality-wise bifurcation of table images in \dataset{} is highlighted in Table \ref{tab:bifurc}. Tables \ref{tab:printed} and \ref{tab:scene} show the script-wise bifurcation along with the distribution of simple tables and complex tables in the printed and scene modalities of \dataset{} respectively. Table \ref{pubtab_dis}, Table \ref{fintab_dis}, and Table \ref{pubtables_dis} describe the distribution of various OTSL cell types in the canonical benchmark datasets \cite{otsl} of PubTabNet, FinTabNet, and PubTables-1M respectively. Similarly, Table \ref{tab:mustard_stats} highlights the statistics of different characters of OTSL vocabulary in the \dataset. Table \ref{tab:dimensions} showcases the average heights and widths of various datasets under consideration.

\renewcommand{\arraystretch}{1.1}
\begin{table}[h]
\centering
\begin{tabular}{|c|c|c|c|}
\hline
\textbf{Modality} & \textbf{Total Size} & \textbf{Simple Tables} & \textbf{Complex Tables} \\ \hline
Scanned and Printed & 1214 & 495 & 719 \\ \hline
Scene               & 214  & 167 & 47  \\ \hline
\textbf{Overall}  & \textbf{1428}       & \textbf{662}           & \textbf{766}            \\ \hline
\end{tabular}
\caption{Statistics of the bifurcation in \dataset{} tables based on modality}
\label{tab:bifurc}
\end{table}

\renewcommand{\arraystretch}{1.1}
\begin{table}[h]
\centering
\begin{tabular}{|c|c|c|c|}
\hline
\textbf{Language} & \textbf{Total Size} & \textbf{Simple Tables} & \textbf{Complex Tables} \\ \hline
Hindi     & 100 & 55 & 45 \\ \hline
Kannada   & 101 & 23 & 78 \\ \hline
Telugu    & 102 & 51 & 51 \\ \hline
Malayalam & 102 & 54 & 48 \\ \hline
Bengali   & 100 & 12 & 88 \\ \hline
Assamese  & 101 & 53 & 48 \\ \hline
Gujarati  & 101 & 35 & 66 \\ \hline
Punjabi   & 102 & 18 & 84 \\ \hline
Tamil     & 100 & 59 & 41 \\ \hline
Oriya     & 101 & 67 & 34 \\ \hline
Urdu      & 102 & 25 & 77 \\ \hline
Chinese   & 102 & 43 & 59 \\ \hline
\textbf{Overall}  & \textbf{1214}       & \textbf{495}           & \textbf{719}            \\ \hline
\end{tabular}
\caption{Overview of scanned and printed tables in \dataset}
\label{tab:printed}
\end{table}

\renewcommand{\arraystretch}{1.1}
\begin{table}[h]
\centering
\begin{tabular}{|c|c|c|c|c|}
\hline
\textbf{Language} & \textbf{Type} & \textbf{Total Size} & \textbf{Simple Tables} & \textbf{Complex Tables} \\ \hline
\multirow{2}{*}{English} & Bordered   & 57 & 41 & 16 \\ \cline{2-5} 
                         & Borderless & 52 & 41 & 11 \\ \hline
\multirow{2}{*}{Chinese} & Bordered   & 54 & 41 & 13 \\ \cline{2-5} 
                         & Borderless & 51 & 44 & 7  \\ \hline
\textbf{Overall}  & \textbf{-}    & \textbf{214}        & \textbf{167}           & \textbf{47}             \\ \hline
\end{tabular}
\caption{Overview of scene tables in \dataset}
\label{tab:scene}
\end{table}

\renewcommand{\arraystretch}{1.1}
\begin{table}[]
\centering
\resizebox{\textwidth}{!}{
\begin{tabular}{|c|c|c|c|c|c|c|}
\hline
{ } & \multicolumn{2}{c|}{{ \textbf{Testset(6942)}}} & \multicolumn{2}{c|}{{ \textbf{Valset(68,002)}}} & \multicolumn{2}{c|}{{ \textbf{Trainset(3,20,000)}}} \\ 
\multirow{-2}{*}{{ \textbf{CHAR}}} & Counts & Avg \% occupancy & Counts  & Avg \% occupancy & Counts     & Avg \% occupancy                                  \\ \hline
F & 436134 & 73.01 & 4145463 & 73.01 & 19545355 & 72.99  \\ \hline
E & 50157  & 5.92  & 486722  & 5.94  & 2302751  & 5.97  \\ \hline
L & 17885  & 2.40  & 163733  & 2.25  & 767628   & 2.21  \\ \hline
X & 8      & 0.00  & 37      & 0.00  & 257      & 0.00  \\ \hline
N & 95748  & 18.58 & 923404  & 18.73 & 4357146  & 18.75  \\ \hline
U & 227    & 0.09  & 1822    & 0.07  & 8617     & 0.07  \\ 
\hline
\end{tabular}
}
\caption{ Coverage of various OTSL cell types in the PubTabNet \cite{pubtabnet} dataset}
\label{pubtab_dis}
\end{table}

\renewcommand{\arraystretch}{1.1}
\begin{table}[]
\centering
\resizebox{\textwidth}{!}{
\begin{tabular}{|c|c|c|c|c|c|c|}
\hline
{ } & \multicolumn{2}{c|}{{ \textbf{Testset(10397)}}} & \multicolumn{2}{c|}{{ \textbf{Valset(10505)}}} & \multicolumn{2}{c|}{{ \textbf{Trainset(88441)}}} \\ 
\multirow{-2}{*}{{ \textbf{CHAR}}} & Counts & Avg \% occupancy & Counts & Avg \% occupancy & Counts & Avg \% occupancy \\ \hline
F & 418006 & 67.00 & 419745  & 68.77 & 4022331  & 67.71 \\ \hline
E & 62262  & 8.25  & 58339   & 7.58  & 651110   & 8.39 \\ \hline
L & 18573  & 3.27  & 15280   & 2.66  & 175672   & 3.07 \\ \hline
X & 0      & 0.00  & 0       & 0.00  & 0        & 0.00 \\ \hline
N & 117955 & 21.47 & 115124  & 20.99 & 1067447  & 20.83 \\ \hline
U & 0      & 0.00  & 0       & 0.00  & 52       & 0.00 \\ 
\hline
\end{tabular}
}
\caption{ Coverage of various OTSL cell types in the FinTabNet \cite{fintabnet} dataset}
\label{fintab_dis}
\end{table}

\renewcommand{\arraystretch}{1.1}
\begin{table}[]
\centering
\resizebox{\textwidth}{!}{
\begin{tabular}{|c|c|c|c|c|c|c|}
\hline
{ } & \multicolumn{2}{c|}{{ \textbf{Testset(92841)}}} & \multicolumn{2}{c|}{{ \textbf{Valset(93989)}}} & \multicolumn{2}{c|}{{ \textbf{Trainset(5,22,874)}}} \\ 
\multirow{-2}{*}{{ \textbf{CHAR}}} & Counts   & Avg \% occupancy  & Counts   & Avg \% occupancy & Counts & Avg \% occupancy \\ \hline
F & 5945261 & 72.80 & 5994435 & 72.78 & 33388077 & 72.76 \\ \hline
E & 349659  & 3.10  & 353354  & 3.08  & 1984485  & 3.08  \\ \hline
L & 384202  & 3.53  & 389780  & 3.52  & 2187423  & 3.54  \\ \hline
X & 758     & 0.01  & 746     & 0.01  & 5017     & 0.01  \\ \hline
N & 1238329 & 18.15 & 1254370 & 18.21 & 6997554  & 18.21 \\ \hline
U & 224747  & 2.40  & 229304  & 2.39  & 1277512  & 2.40 \\ 
\hline
\end{tabular}
}
\caption{ Coverage of various OTSL cell types in the PubTables-1M \cite{tatr-pub-1m} dataset}
\label{pubtables_dis}
\end{table}

\renewcommand{\arraystretch}{1.1}
\begin{table}[]
\centering
\begin{tabular}{|c|cc|cc|}
\hline
\multirow{2}{*}{\textbf{CHAR}} & \multicolumn{2}{c|}{\textbf{Scanned or Printed Tables(1214)}} & \multicolumn{2}{c|}{\textbf{Scene Text Tables (214)}}   \\ \cline{2-5} & \multicolumn{1}{c|}{Counts} & Avg \% occupancy & \multicolumn{1}{c|}{Counts} & Avg \% occupancy \\ \hline
F & \multicolumn{1}{c|}{47357} & 70.82 & \multicolumn{1}{c|}{3695} & 70.71 \\ \hline
E & \multicolumn{1}{c|}{1854}  & 2.51  & \multicolumn{1}{c|}{545}  & 7.40  \\ \hline
L & \multicolumn{1}{c|}{4429}  & 5.89  & \multicolumn{1}{c|}{120}  & 1.48  \\ \hline
X & \multicolumn{1}{c|}{17}    & 0.02  & \multicolumn{1}{c|}{1}    & 0.01  \\ \hline
N & \multicolumn{1}{c|}{11094} & 19.52 & \multicolumn{1}{c|}{1175} & 22.63 \\ \hline
U & \multicolumn{1}{c|}{1155}  & 1.24  & \multicolumn{1}{c|}{115}  & 1.23  \\ \hline
\end{tabular}
\caption{ Coverage of various OTSL cell types for \dataset}
\label{tab:mustard_stats}
\end{table}

\renewcommand{\arraystretch}{1.1}
\begin{table}[]
\centering
\begin{tabular}{|c|c|c|c|}
\hline
\textbf{Dataset}                                                                              & \textbf{Split} & \textbf{Avg Width} & \textbf{Avg Height} \\ \hline
\multirow{3}{*}{\textbf{PubTabNet}} & Train & 405  & 210 \\ \cline{2-4} 
                                    & Val   & 399  & 207 \\ \cline{2-4} 
                                    & Test  & 406  & 213 \\ \hline
\multirow{3}{*}{\textbf{FinTabNet}} & Train & 440  & 180 \\ \cline{2-4} 
                                    & Val   & 446  & 173 \\ \cline{2-4} 
                                    & Test  & 438  & 176 \\ \hline
\multirow{3}{*}{\textbf{\begin{tabular}[c]{@{}c@{}}PubTables-1M\\ (5k samples)\end{tabular}}} & Train          & 579                & 315                 \\ \cline{2-4} 
                                    & Val   & 578  & 314 \\ \cline{2-4} 
                                    & Test  & 579  & 316 \\ \hline
\textbf{\dataset}                    & Test  & 1450 & 776 \\ \hline
\end{tabular}
\caption{Dimesnions of images for all the datasets within our experimentation}
\label{tab:dimensions}
\end{table}

\section{Role of Table Grid Estimator}
\begin{figure}[h]
  \centering
  \begin{subfigure}[b]{\textwidth}
    \includegraphics[width=\textwidth]{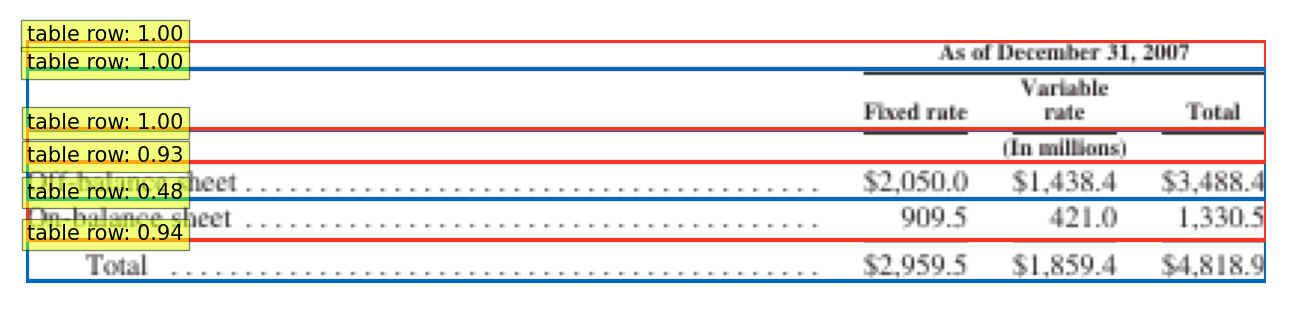}
    \caption{Estimating the number of rows}
    \label{fig:sphy1}
  \end{subfigure}
  %\quad % Add some space between the subfigures
  \begin{subfigure}[b]{\textwidth}
    \includegraphics[width=\textwidth]{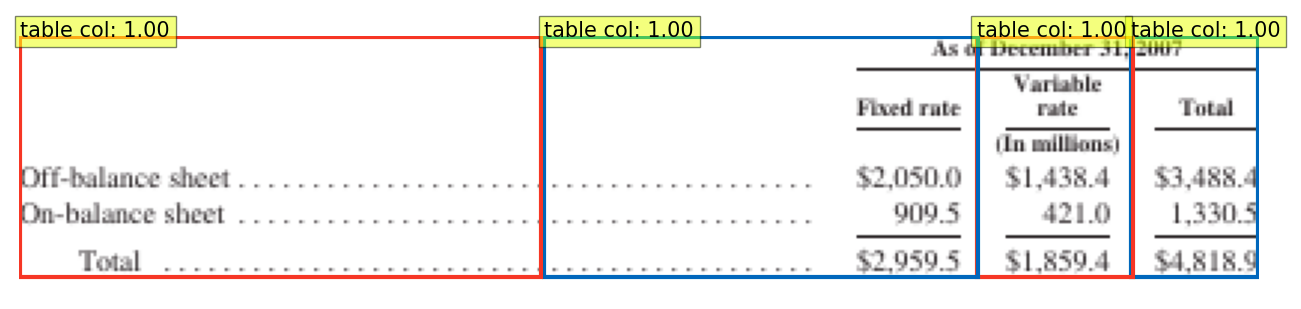}
    \caption{Estimating the number of columns}
    \label{fig:phy2}
  \end{subfigure}
  \caption{Predictions for a table image from FinTabNet dataset using TATR v1.1}
  \label{fig:phy}
\end{figure}

We have used the TATR \cite{aligning} V1.1 model pre-trained on FinTabNet, PubTabNet, and PubTables-1M datasets for determining the physical structure of the input table image. The underlying architecture for TATR is DETR \cite{detr} which is trained for six classes including 'table-row' and 'table-column'. For grid alignment, we make use of classes of 'table-row' and 'table-column' to determine the number of rows (\row) and columns (\col). We do not use the other four classes: 'table', 'table column header', 'table projected row header', and 'table spanning cell' classes. The 'table' class is unused as it is not useful for independent TSR tasks as the TSR system assumes the detected table image as input. The rest of the classes always overlap with some of the rows and columns covered in the 'table-row' and 'table-column' classes and add no value to retrieve the count of table rows and columns. Hence we only use the 'table-row' and 'table-column' classes. and We use a detection threshold of $0.25$  during the inference of the TATR V1.1 model. We further carry out NMS for table row class with an IOU threshold of $0.25$ to minimize overlapping predictions and improve the row match. Once \row{} and \col{} are determined, these values align with the OTSL sequence predicted by \name. Fig \ref{fig:phy} shows the output of the TATR model for detecting table rows and table columns respectively.

\section{Role of \name{} to get OTSL Predictions}
\begin{figure}
    \centering
    \makebox[0pt]{\includegraphics[width=\textwidth]{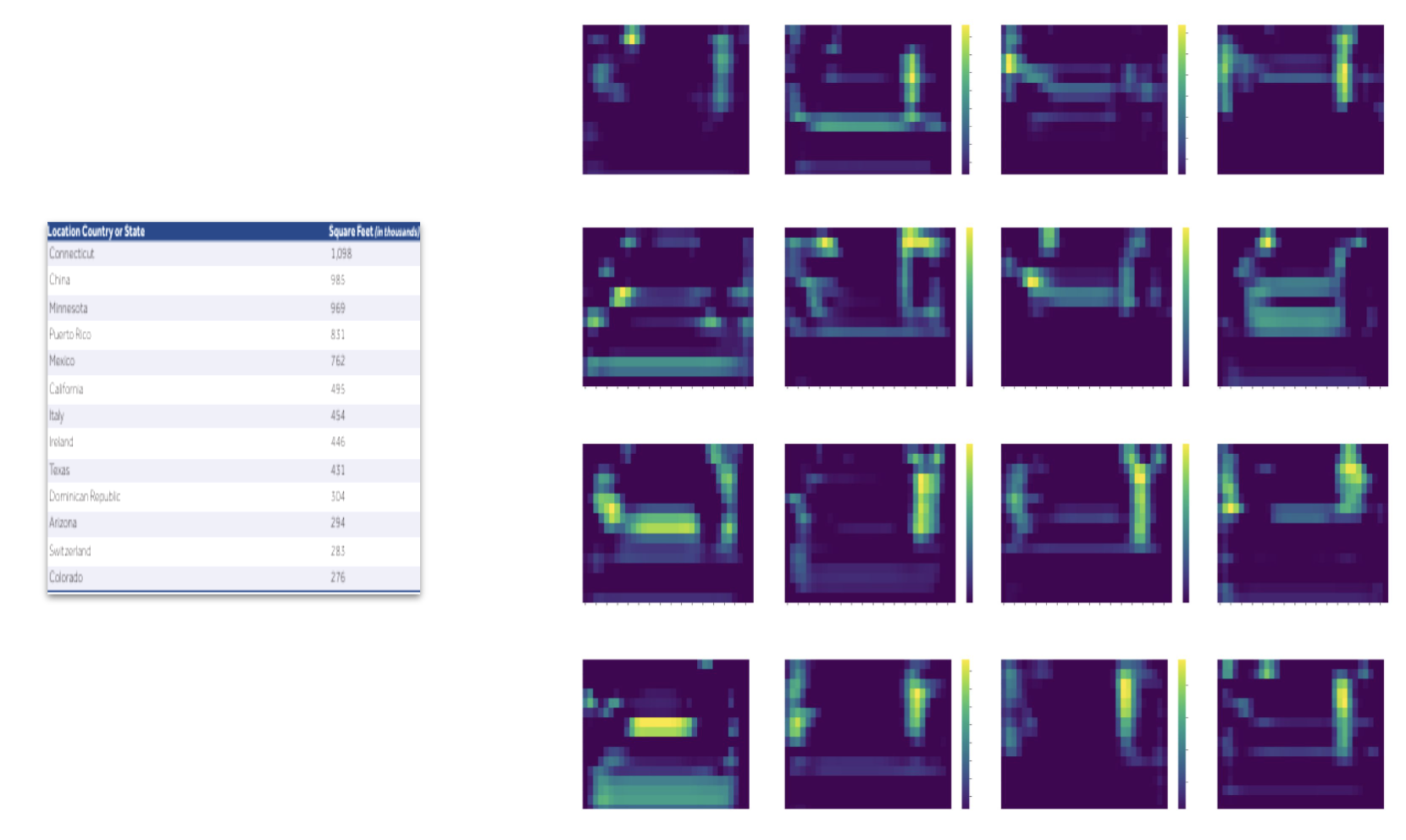}}
    \caption{Illustration of random 16 feature maps, each of size $16 * 32$, generated over a preprocessed table image (left) of size $128 * 128$ using GCA-based encoder in \name. The feature tensor is encoded and later sent to the decoder} 
    \label{fig:features_img}
\end{figure}

\begin{figure}
    \centering
    \makebox[0pt]{\includegraphics[scale=0.38]{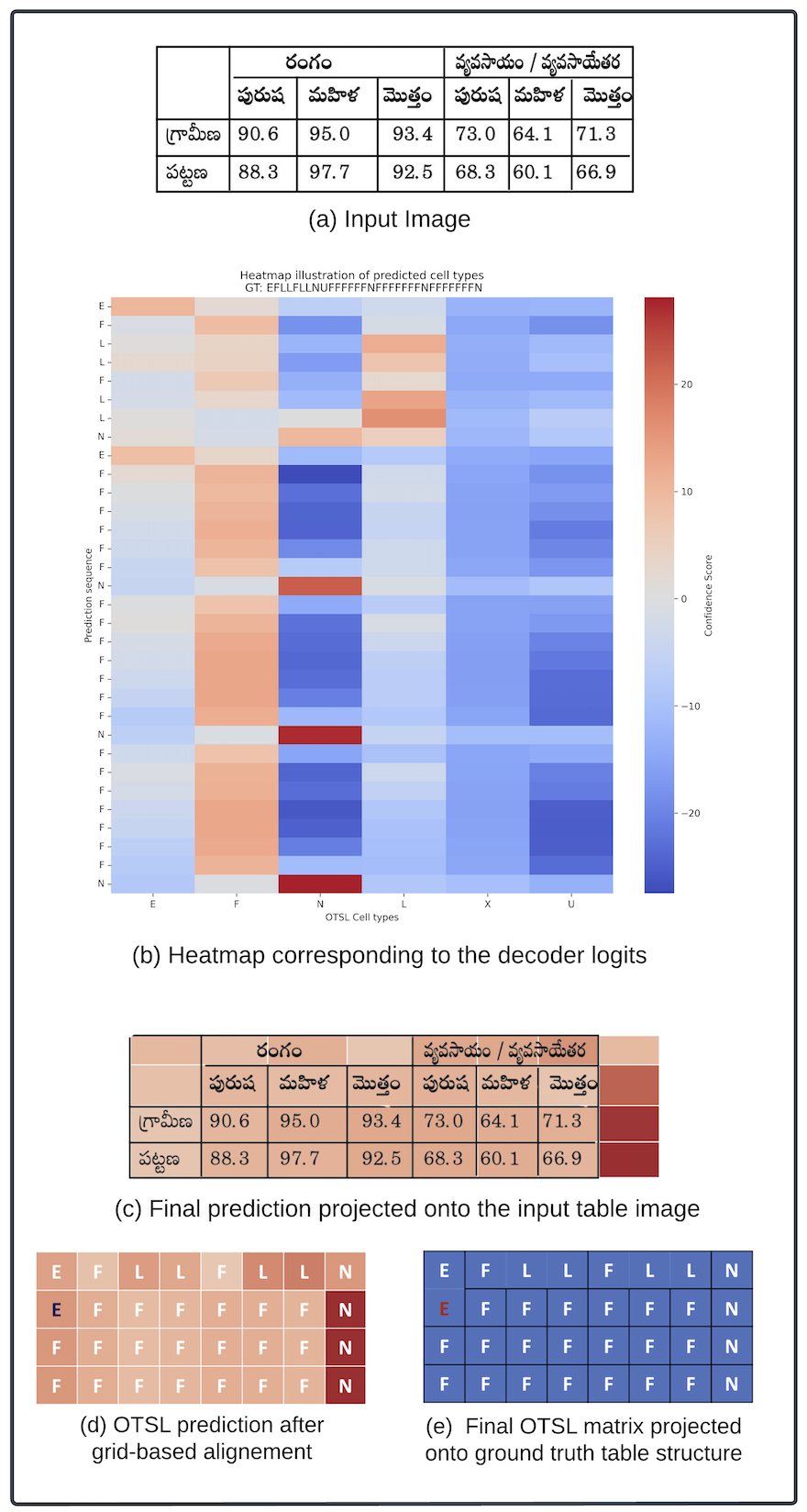}}
    \caption{Illustration of decoding table structure using \name} 
    \label{fig:illustration}
\end{figure}

\name{} is composed of a GCA-based encoder and a decoder capable of producing an OTSL sequence. All images fed into \name{} encoder are downsampled to a size of $128 * 128$ to view this as a script-agnostic arrangement of cells. Fig \ref{fig:features_img} shows 16 random feature maps out of the 512 channels. This feature tensor undergoes positional embedding to finally produce encoded vectors that are passed to the \name{} decoder to predict the OTSL string. An illustration from Fig \ref{fig:illustration} shows the functioning of the decoder. As shown in subfig (a), the input table image is preprocessed and passed through the GCA-based encoder to produce an encoded feature tensor. These features are passed to the \name{} decoder which is composed of multi-head cross attention, masked multi-head attention, and feed-forward neural network \cite{master}. The decoder produces logits as seen in subfig (b). Every n$^{th}$ row of the logits represents a probability distribution of the n$^{th}$ predicted character corresponding to one of the six characters (as shown in the columns of the logits) from the OTSL vocabulary. The OTSL sequence results from the concatenation of every character for which the probability from the particular row is maximum. The final predicted OTSL string along with its confidence score is projected onto the table in subfig (c) where the table grid estimator gives us four rows and seven columns. Subfig (d) highlights the $4 * 8$ OTSL matrix obtained after the grid-based alignment. Finally, subfig (e) showcases the final OTSL matrix projected onto the ground truth HTML table structure where the red entries in the matrix denote an incorrect prediction. 

\section{Ablation Study}
We report the results of several experiments using different parameters of \name{} for popular benchmark TSR datasets that include PubTabNet\cite{pubtabnet}, FinTabNet\cite{fintabnet}, and PubTables-1M\cite{tatr-pub-1m}. Table \ref{tab:ablation} highlights the performance of different experiments defined by different configurations that include the number of decoder layers and shape (dimensions) of preprocessed images. Because the majority of the OTSL sequences have lengths under 224, we use the maximum permissible length of 224 for all our experiments. PubTabNet and PubTables-1M datasets perform well when the model is trained on all merged data. However, the FinTabNet test set shows marginally better scores for a model that is only trained on FinTabNet training data. We report the best-performing results for the individual datasets in the main paper.

\renewcommand{\arraystretch}{1.2}
\begin{table}[]
\centering
\resizebox{\textwidth}{!}{
\begin{tabular}{|c|c|c|c|c|c|}
\hline
\textbf{Test} &
  \textbf{Training} &
  \textbf{Config} &
  \textbf{\begin{tabular}[c]{@{}c@{}}TEDS-S \\ Simple\end{tabular}} &
  \textbf{\begin{tabular}[c]{@{}c@{}}TEDS-S \\ Complex\end{tabular}} &
  \textbf{\begin{tabular}[c]{@{}c@{}}TEDS-S \\ Overall\end{tabular}} \\ \hline
\multirow{4}{*}{PubTabNet}    & PubTabNet    & *Layers: 3, Shape: 32*128  & 97.91          & 91.17          & 94.61          \\ \cline{2-6} 
                              & PubTabNet    & Layers: 4, Shape: 32*128  & 98.12          & 92.84          & 95.53          \\ \cline{2-6} 
                              & All          & Layers: 6, Shape: 32*128  & \textbf{98.11} & 92.98          & 95.60          \\ \cline{2-6} 
                              & All          & Layers: 6, Shape: 128*128 & 98.00          & \textbf{93.32} & \textbf{95.71} \\ \hline
\multirow{4}{*}{FinTabNet}    & FinTabNet    & Layers: 6, Shape: 32*32   & \textbf{98.39} & 94.57          & 96.41          \\ \cline{2-6} 
                              & FinTabNet    & Layers: 6, Shape: 32*128  & 98.30          & 97.46          & 97.88          \\ \cline{2-6} 
                              & All          & Layers: 6, Shape: 128*128 & 98.31          & 97.73          & 98.01          \\ \cline{2-6} 
                              & FinTabNet    & Layers: 6, Shape: 128*128 & 98.35          & \textbf{97.74} & \textbf{98.03} \\ \hline
\multirow{4}{*}{PubTables-1M} & PubTables-1M & Layers: 6, Shape: 32*128  & 98.19          & 92.69          & 95.50          \\ \cline{2-6} 
                              & All          & Layers: 8, Shape: 32*128  & 98.88          & 93.34          & 96.00          \\ \cline{2-6} 
                              & All          & Layers: 6, Shape: 32*128  & 98.87          & 94.80          & 96.75          \\ \cline{2-6} 
                              & All          & Layers: 6, Shape: 128*128 & \textbf{98.92} & \textbf{96.54} & \textbf{97.68} \\ \hline
\end{tabular}
}
\caption{Results on different test sets for \name{} trained on various datasets. The training set of 'All' refers to the combined training dataset of all three datasets. The config is dictated by two parameters mainly the number of decoder layers and shape (dimensions) the input image is resized to in the preprocessing stage. * indicates that the maximum permissible length of prediction was set to 192 for that experiment}
\label{tab:ablation}
\end{table}

\section{Inference Time Calculation}
As explained in the methodology section, it is evident that the total time for logical structure prediction is the sum of the duration of each of the four concerned steps. The total time ($T_{\text{total}}$) can be represented as the sum of the individual times:
\begin{equation}
T_{\text{total}} = T_{\text{\name}} + T_{\text{Post-Processing}} 
\end{equation}
\begin{equation}
T_{\text{Post-Processing}} = T_{\text{Grid}} + T_{\text{Alignment}} + T_{\text{Conversion}}
\end{equation}
Where:
\begin{itemize}
  \item $T_{\text{\name}}$: SPRINT Prediction time to get OTSL sequence
  \item $T_{\text{Post-Processing}}$: Sum of the times for steps following \name{} prediction
  \item $T_{\text{Grid}}$: Grid Estimation time using TATR\cite{tatr-pub-1m}
  \item $T_{\text{Alignment}}$: Grid-based alignment time
  \item $T_{\text{Conversion}}$: OTSL to HTML Conversion time
\end{itemize}

Our experiments of determining inference time involved determining $T_{\text{total}}$ as the sum of $T_{\text{\name}}$ and $T_{\text{Post-Processing}}$. We curated a random set of 400 images from FinTabNet \cite{fintabnet} and calculated the  $T_{\text{\name}}$ and $T_{\text{Post-Processing}}$ as the average running time for 50 iterations. We estimated $T_{\text{\name}}$ and $T_{\text{Post-Processing}}$ as 1.36 seconds and 0.16 seconds respectively. Thus, for our approach, the total time comes out to be 1.52 seconds. We performed the same experiment to determine the inference time for MTL-TabNet \cite{mtl-tab-net} for a structure decoder that directly gives an HTML sequence as output. All the experiments were performed on NVIDIA RTX A6000 single GPU. Table \ref{tab:inftime} highlights the different approaches and their average inference time for table structure predictions. Since Tableformer \cite{tableformer} does not provide its code and checkpoints, we use the pre-reported average prediction times \cite{otsl} for the entries of \textit{TableFormer + OTSL} and \textit{TableFormer + HTML}. The same statistics are used in the main paper to present the comparative analysis of \name{} with other approaches. As seen in Table \ref{tab:inftime}, our approach is the fastest of all the enlisted approaches. 

\renewcommand{\arraystretch}{1.1}
\begin{table}[]
\centering
\resizebox{\textwidth}{!}{
\begin{tabular}{|c|c|c|c|}
\hline
\textbf{Approach}  & \textbf{Vocabulary} & \textbf{Prediction Time (sec)} & \textbf{Hardware}                                                                       \\ \hline
TableFormer + HTML \cite{tableformer} & HTML                & 3.26                               & \multirow{2}{*}{\begin{tabular}[c]{@{}c@{}}AMD EPYC \\ 7763 CPU\end{tabular}}           \\ \cline{1-3}
TableFormer + OTSL \cite{otsl} & OTSL & 1.85        &  \\ \hline
MTL-TabNet \cite{mtl-tab-net}         & HTML                & 2.35                               & \multirow{2}{*}{\begin{tabular}[c]{@{}c@{}}Single NVIDIA RTX \\ A6000 GPU\end{tabular}} \\ \cline{1-3}
\textbf{Ours}      & OTSL & \textbf{1.52} &  \\ \hline
\end{tabular}
}
\caption{Comparative overview of inference time of different approaches per image for predicting table structures. We use the pre-reported average prediction times for TableFromer \cite{tableformer} as mentioned in the OTSL baseline \cite{otsl} results}
\label{tab:inftime}
\end{table}

\bibliographystyle{splncs04}
\bibliography{main}

\end{document}